\documentclass[10pt,twocolumn,letterpaper]{article}

\usepackage{cvpr}              % To produce the CAMERA-READY version
\usepackage[accsupp]{axessibility}  % Improves PDF readability for those with disabilities.

\definecolor{cvprblue}{rgb}{0.21,0.49,0.74}
\usepackage[pagebackref,breaklinks,colorlinks,allcolors=cvprblue]{hyperref}

\def\paperID{} % *** Enter the Paper ID here
\def\confName{CVPR}
\def\confYear{2026}

\usepackage{amsfonts}
\usepackage{mathtools}
\usepackage{amssymb}
\usepackage{amsmath}

\usepackage{xcolor}
\usepackage{multirow}
\usepackage{colortbl}
\usepackage{adjustbox}
\usepackage{tabularx, booktabs}
\usepackage{caption}
\usepackage{tcolorbox}
\usepackage{enumitem}
\usepackage{booktabs}
\usepackage{siunitx}
\usepackage{wrapfig}

\usepackage{siunitx}
\newcommand{\add}[1]{{\color{blue} #1}}
\newcommand{\hj}[1]{{\color{cyan} HJ: #1}}
\newcommand{\jh}[1]{{\color{purple} JH: #1}}

\title{VIRO: Robust and Efficient Neuro-Symbolic Reasoning with Verification for Referring Expression Comprehension}

\author{
Hyejin Park
\hfill
Junhyuk Kwon
\hfill
Suha Kwak
\hfill
Jungseul Ok$^{\dagger}$\\
Pohang University of Science and Technology (POSTECH), South Korea\\
{\tt\small \{parkebbi2, treejhk, suha.kwak, jungseul\}@postech.ac.kr}
}

\begin{document}
\maketitle
\def\thefootnote{}\footnotetext{{$^\dagger$}: corresponding author}

\begin{abstract}
% Background – However… (연구할 분야의 문제점) – 그래서~ 우리 논문에서 한 것(목적성) – 특히~(구체성) – (연구결과의 해석) 

Referring Expression Comprehension (REC) aims to localize the image region corresponding to a natural language query. 
Recent neuro-symbolic REC approaches leverage large language models (LLMs) and vision-language models (VLMs) to perform compositional reasoning, decomposing queries into structured programs and executing them step-by-step. While such approaches achieve interpretable reasoning and strong zero-shot generalization, they assume that intermediate reasoning steps are accurate.
However, this assumption causes cascading errors: false detections and invalid relations propagate through the reasoning chain, yielding high-confidence false positives even when no target is present in the image. To address this limitation, we introduce Verification-Integrated Reasoning Operators (VIRO), a neuro-symbolic framework that embeds lightweight operator-level verifiers within reasoning steps.
Each operator executes and validates its output, such as object existence or spatial relationships, allowing the system to robustly handle no-target cases through verification-aware abstention.
Our framework achieves state-of-the-art performance, reaching 61.1\% balanced accuracy across target-present and no-target settings, and demonstrates generalization to real-world egocentric data. VIRO also shows high reliability with a program failure rate of at most 0.3\%, efficient per-query runtime, and scalability through decoupled program generation and execution. Code is available at \href{https://github.com/ml-postech/VIRO-neuro-symbolic-reasoning-with-verification}{https://github.com/ml-postech/VIRO-neuro-symbolic-reasoning-with-verification}.

\end{abstract}
\section{Introduction}

% Introduction to Referring Expression Comprehension (REC)
Humans naturally interpret visual scenes through structured language, 
such as ``the blue cup on the wooden table," to identify objects through a sequence of semantic relations.
This fundamental human skill is formalized in the vision-language task of Referring Expression Comprehension (REC), where the goal is to localize a target object in an image based on a natural language description \citep{qiao2020referring}. This task has broad applicability, including vision-language navigation \citep{wang2021structured, zhang2024interactive}, human-robot interaction \citep{shridhar2022cliport, jin2025referring}, and text-to-image retrieval~\citep{lee2024interactive}.
% However, most existing systems are forced to output a bounding box even when no object in the scene matches the query, leading to hallucinated detections in no-target scenarios.

\begin{figure*}
  \centering
  \includegraphics[width=0.78\linewidth,trim=1 0 1 0,clip]{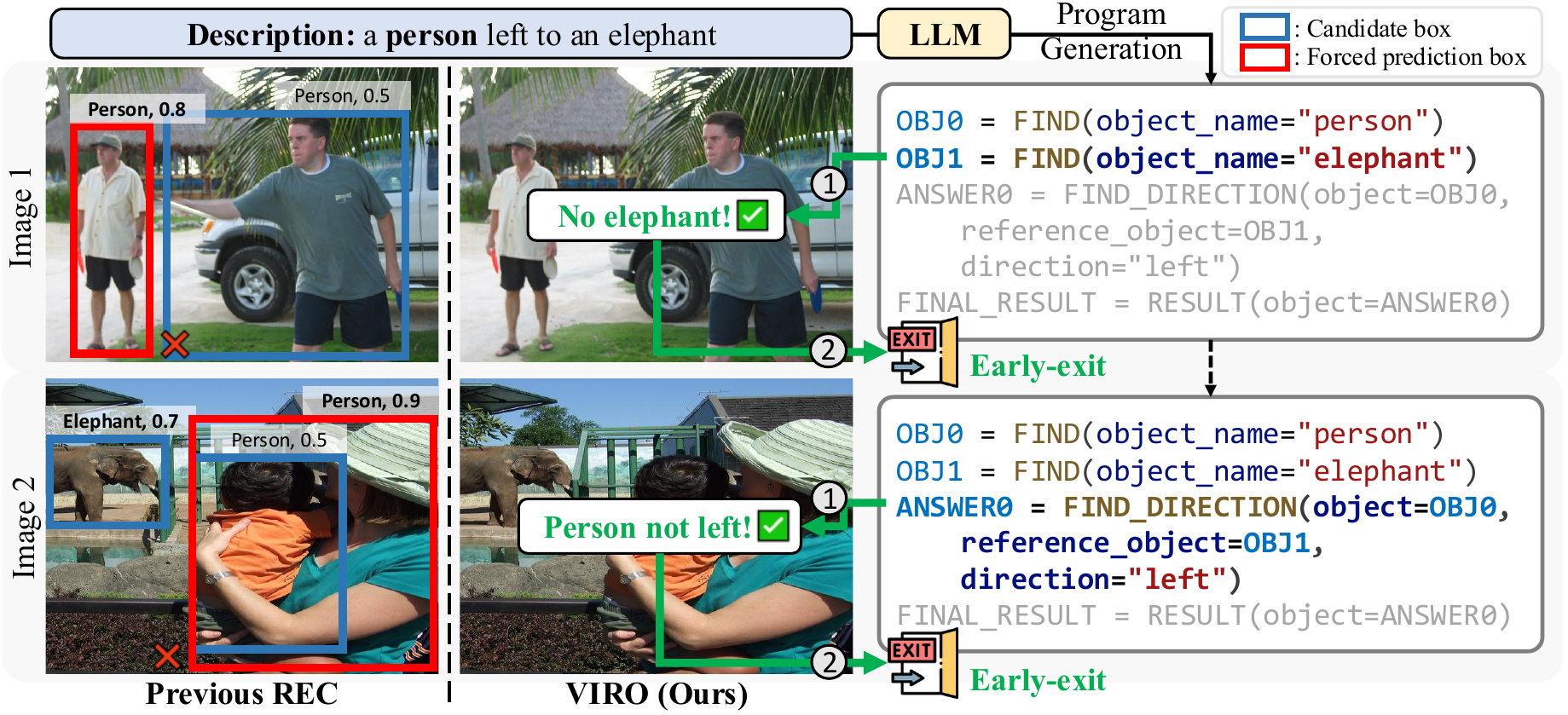}
  \vspace{-.5em}
  \caption{\textbf{Missing verification leads to forced predictions, which VIRO prevents.} 
  Illustrative comparison between previous REC methods and our VIRO framework in no-target cases. 
  Previous REC methods (left) are forced to output a prediction, even when the query cannot be grounded in the image, due to the lack of a mechanism for eliminating incorrect candidates. 
  In contrast, our VIRO framework (right) terminates early instead of hallucinating a prediction: 
  (i) \texttt{FIND} operator identifies that there is no elephant in the image (top); (ii) \texttt{FIND\_DIRECTION} operator identifies the person is not positioned to the left of the elephant (bottom).}
  \label{fig:viro}
  \vspace{-1.3em}
\end{figure*}

Early REC approaches relied on supervised end-to-end models that directly map textual queries to image regions~\citep{yu2018mattnet, kamath2021mdetr, yan2023universal}, which lack generalization. More recent work has moved toward compositional and LLM-driven reasoning, where a query is parsed into a program or sequence of modular operators for object detection, attribute filtering, and spatial reasoning~\citep{subramanian2022reclip, han2024zero, shen2024groundvlp}. These methods often build on open-vocabulary detectors (OVDs)~\citep{li2022grounded, liu2024grounding, xiao2024florence} to align visual candidates with natural-language cues and generalize to unseen compositions~\citep{suris2023vipergpt, ke2024hydra, cai2025naver, chen2025flora}. While improving interpretability, these methods still lack verification of intermediate reasoning steps.

In existing compositional pipelines, intermediate outputs, such as object proposals, attribute matches, and spatial relations, are typically assumed to be correct and are propagated without explicit checks. This leaves the system vulnerable to cascading errors, especially when OVDs produce high-confidence false positives for non-existent objects~\citep{li2023evaluating}. The problem is particularly severe in no-target cases, where pipelines are effectively forced to select one of these false positives as the answer. 
Moreover, many of these systems suffer from efficiency and scalability issues: (i) existing systems often place heavy multimodal LLMs in the inner loop, which leads to substantial latency, and (ii) program generation is tightly coupled with execution, often requiring regeneration of a separate program for each frame in a video, causing computational cost to grow linearly.

To address these challenges, we propose Verification-Integrated Reasoning Operators (VIRO), a neuro-symbolic framework that integrates verification at the operator level.
Our pipeline first uses an LLM to translate a natural-language description into a sequence of executable operators, and then executes them sequentially, as illustrated in Figure~\ref{fig:viro}.
Each applicable operator not only performs a reasoning action but also validates its own result.
These include uncertainty verification, where lightweight CLIP-based filters suppress high-confidence false positives from OVDs, and logical verification, which checks whether spatial and relational constraints are actually satisfied.
When verification does not pass, indicating that no valid target exists, the operator abstains and triggers early termination, enabling explicit no-target detection.% rather than forcing a hallucinated prediction.

% By transforming compositional reasoning into a self-verifying process, VIRO yields interpretable, efficient, and robust visual reasoning.

% Crucially, VIRO embeds verification within each reasoning step, allowing operators to abstain from forced predictions and to terminate early when conditions are not met. This verification process includes two key components: a lightweight CLIP-based filter with minimal computational overhead that suppresses high-confidence false positives from open-vocabulary detectors, and logical checks that strictly enforce spatial and relational constraints, yielding robustness in no-target scenarios.

% The operator-level verification is tailored to each function. For example, the \texttt{FIND} operator uses an OVD to output candidate regions, which can sometimes produce high-confidence false positives. 
% To address this, we incorporate a CLIP-based uncertainty filter within the \texttt{FIND} operator, which filters out spurious detections with minimal computational overhead.
% For relational operators, such as \texttt{FIND\_DIRECTION}, which assess the relationship between objects (e.g., `person' and `elephant' in Figure~\ref{fig:viro_comp}), we introduce logical verification to validate geometric relationships, ensuring that spatial constraints are strictly satisfied.
% By embedding verification directly within each reasoning step, VIRO can reliably reject invalid candidates, terminate early if empty, and handle no-target scenarios in an interpretable and efficient manner.

We evaluate VIRO in a zero-shot setting on both no-target and standard REC scenarios. To highlight the importance of verification, we first demonstrate strong robustness on the gRefCOCO no-target dataset~\citep{he2023grec, liu2023gres}, where images contain no instance matching the query. Furthermore, VIRO achieves state-of-the-art performance compared to compositional baselines on standard REC benchmarks, including RefCOCO/+/g~\citep{yu2016modeling, mao2016generation} and the adversarial RefAdv dataset~\citep{akula2020words}. 
Beyond accuracy, VIRO achieves low program failure rates and efficient per-query runtime. By amortizing a single generated program across multiple images, it scales seamlessly to the $1$-query--$N$-images setting, which we comprehensively validate on the egocentric RefEgo dataset~\citep{kurita2023refego}---a benchmark comprising both target-present and no-target frames.
% and scales efficiently to the $1$-query--$N$-images setting by amortizing a single program across multiple images. Finally, we demonstrate the extensibility of our approach to real-world scenarios through egocentric experiments on the RefEgo dataset~\citep{kurita2023refego}

%Our framework is evaluated on both no-target and standard REC scenarios. On the gRefCOCO no-target dataset~\citep{he2023grec}, it exhibits strong robustness, achieving state-of-the-art accuracy when compared to compositional baselines across standard REC benchmarks, including RefCOCO/+/g.
%Furthermore, our neuro-symbolic approach achieves high throughput (FPS), a low program failure rate, and enhanced scalability for processing multiple images from a single query.

% Our approach also achieves high throughput (FPS) with the design choices of the early-exit mechanism and lightweight CLIP filter. Additionally, VIRO shows a low program failure rate of just 0.3\%, significantly outperforming other compositional reasoning methods, which have failure rates at least above 5\%.
% Finally, our neuro-symbolic design separates pre-execution for program generation and execution time, offering improved scalability for processing multiple images from a single query.

% operators (CLIP-based..)
% framework (early-exit)
% show ...
Our key contributions are summarized as follows:
\begin{itemize}[leftmargin=10pt, itemsep=0pt, topsep=0pt]
    \item We identify forced prediction as a fundamental failure mode of compositional vision-language reasoning, and introduce verification-integrated operators to address it.
    % \item We propose VIRO, a neuro-symbolic REC framework that embeds lightweight verification into each reasoning step via operator-level programs, enabling robust execution with a very low program failure rate.

    \item We design operator-level programs in which each operator both acts and verifies, using uncertainty- and logic-based checks to suppress error cascades and enable explicit no-target detection.

    % \item Each operator both acts and verifies, using uncertainty- and logic-based checks to suppress error cascades and perform efficient, explicit no-target detection, resulting in high reliability and low per-query execution cost.
    % \item Each program operator both acts and verifies, using uncertainty- and logic-based checks to suppress error cascades and enable efficient, explicit no-target detection.
    % \item Each program operator both acts and verifies, using uncertainty- and logic-based checks to suppress error cascades and to perform efficient, explicit no-target detection, resulting in high execution throughput.

    \item In a zero-shot setting, VIRO demonstrates strong robustness on the gRefCOCO no-target split and achieves state-of-the-art performance among compositional baselines on standard REC benchmarks. 
    By decoupling program synthesis from execution and amortizing a single program over many images, VIRO scales favorably in the $1$-query--$N$-image regime.% and extends to real-world egocentric data through experiments on RefEgo.
    % By decoupling program synthesis from execution and amortizing a single program over many images, VIRO scales favorably in the $1$-query-$N$-image regime and demonstrates extensibility to real-world applications on the egocentric RefEgo dataset.
\end{itemize}

\begin{table*}[tb!]
\centering
\caption{\textbf{Overview of VROs}. All operators are designed to return a set of verified bounding boxes or an empty set ($\varnothing$) if its condition is not satisfied. Additional operators are provided in Appendix~\ref{appendix:operators}.}
\vspace{-.5em}
\begin{adjustbox}{width=.8\linewidth}
\begin{tabular}{llll}
\toprule
\textbf{Operator} & \textbf{Input Arguments} & \textbf{Verification Module} & \textbf{Built-in Models} \\
\midrule
\texttt{FIND} & \texttt{object\_name} & CLIP-based verifier & OVD, CLIP \\
\texttt{PROPERTY} & \texttt{object}, \texttt{attribute} & CLIP score  & CLIP\\
% \texttt{FIND\_DIRECTION} & \texttt{object}, \texttt{reference\_object}, \texttt{direction}& Directional geometric predicate & - \\
\texttt{RELATIVE\_DEPTH} & \texttt{object}, \texttt{reference\_object}, \texttt{criteria} & Relative depth relation & DepthAnything \\
\bottomrule
\vspace{-2.5em}
\end{tabular}
\end{adjustbox}
\label{tab:operator}
\end{table*}

\section{Related Work}
\paragraph{Referring Expression Comprehension.}
Early end-to-end REC models~\citep{yan2023universal,kamath2021mdetr} achieve strong performance on seen domains but suffer under distribution shifts.
To improve zero-shot generalization, recent work has moved toward first decomposing the textual description into \textit{structured} semantic units and then aligning them with candidate proposals generated by Faster R-CNN~\citep{ren2016faster}, as in MAttNet~\citep{yu2018mattnet}.
Building on this paradigm, ReCLIP~\cite{subramanian2022reclip} combines CLIP-based matching with rule-based spatial reasoning, SS-CLIP~\cite{han2024zero} aligns textual and visual triplets via structural similarity, GroundVLP~\citep{shen2024groundvlp} fuses heatmaps derived from vision–language pre-training models with proposals from an open-vocabulary detector, and FLORA~\cite{chen2025flora} formalizes natural-language queries into a formal language for probabilistic matching.
While these approaches enable zero-shot transfer, they remain inflexible when queries deviate from pre-defined forms, limiting their ability to handle diverse linguistic inputs.
% This motivates a line of work that leverages neuro-symbolic program-based reasoning to achieve more flexible compositions for REC.
\vspace{-.5em}
\paragraph{Compositional Reasoning REC.}
Compositional approaches aim to provide more flexible reasoning than fixed structured pipelines by decomposing a query into interpretable operations.
ViperGPT~\citep{suris2023vipergpt} demonstrates this idea by generating executable Python code, but its reliance on unconstrained free-form code generation often leads to non-runnable programs. To improve reliability, we follow operator-based generation frameworks similar to VisProg~\citep{gupta2023visual}, which restrict the program space to well-defined primitives.
HYDRA~\citep{ke2024hydra} introduces iterative planning with a planner–reasoner loop, and NAVER~\citep{cai2025naver} performs self-correcting inference over multiple reasoning states, with a final verification step to check whether the predicted target matches the query.
However, these approaches rely heavily on large multimodal models, \textit{e.g.}, BLIP-2~\citep{li2023blip}, InternVL2~\citep{chen2024far}, which offers limited protection against execution-time reasoning errors and incurs substantial overhead.
Moreover, existing methods typically assume that a valid target always exists, often forcing a prediction even when no object actually matches the query.
In contrast, our framework integrates lightweight verification at the operator level to robustly reject no-target cases.

\section{Method}\label{sec:method}

% We propose PREVO, Pre-execution REasoning with Verification-integrated Operators, a decoupled neuro-symbolic framework designed for robust and efficient Referring Expression Comprehension (REC).
In Section~\ref{sec:method:formulation}, we formally define the REC problem, extended to handle no-target cases. Subsequently, we present the neuro-symbolic reasoning pipeline VIRO in Section~\ref{sec:method:pipeline}. % introduces our VIRO, followed by the pre-execution stage in Section~\ref{sec:method:pre-exe} and the execution stage in Section~\ref{sec:method:exec}.

\subsection{Problem Formalization}\label{sec:method:formulation}

REC aims to localize a region within an image $I$ that corresponds to a given natural language query $Q$.
Conventional REC assumes that the target object described by the query is always present in the image.
This assumption does not hold in practical applications, such as a visual search system or a robot searching for an object in a building \citep{yokoyama2024vlfm, zhang2024interactive, yin2025unigoal}, where the target is frequently absent from most images.
We formalize the output of the model as:
\begin{align}
    Y =
    \begin{cases}
    B, & \text{if a target exists in } I \;, \\
    \varnothing, & \text{otherwise} \;.
    \end{cases}
\end{align}
Here, $B=(x, y, w, h)\in\mathbb{R}^4$ denotes a bounding box in pixel coordinates, where $(x, y)$ represents the center coordinates of the box, and $w, h$ are its width and height. The $\varnothing$ denotes the absence of a target,  \textit{i.e.}, no object corresponding to the query $Q$ is present in the image.

\subsection{Verification-Integrated Reasoning Operators}\label{sec:method:pipeline}

\begin{figure*}
  \centering
  \includegraphics[width=0.82\linewidth,trim=1 0 1 0,clip]{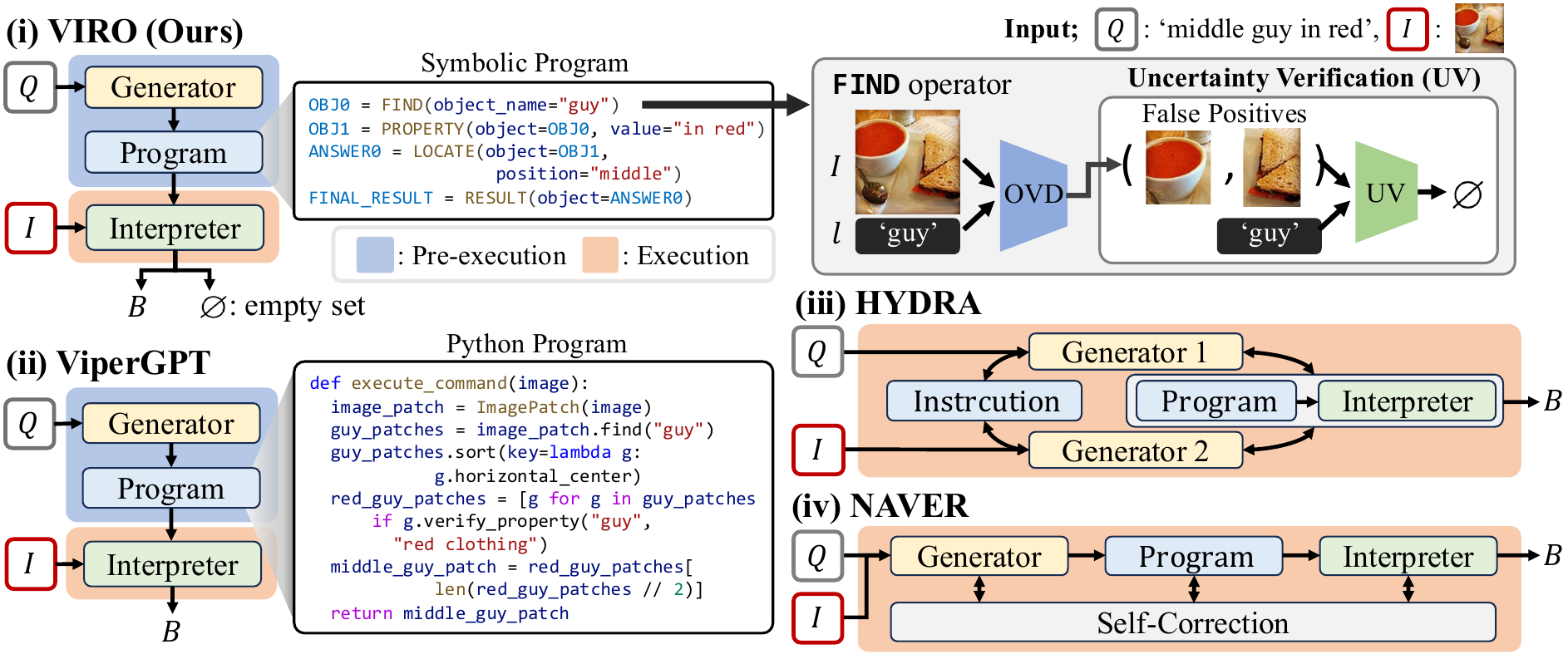}
  \vspace{-.3em}
  \caption{\textbf{VIRO vs. prior compositional REC.} 
  (i) VIRO generates a symbolic program with an LLM, then an interpreter executes operators with \emph{operator-level verification}; as exemplified by the \texttt{FIND} operator using CLIP-based uncertainty verification to prune OVD proposals.
  In contrast, (ii) ViperGPT compiles queries into Python code but lacks verification, while (iii) HYDRA and (iv) NAVER tightly couple program generation and execution, requiring iterative generators for all inputs.}
  \vspace{-1.2em}
  \label{fig:viro_comp}
\end{figure*}

To address the task defined above, VIRO employs a two-stage neuro-symbolic pipeline. As illustrated in Figure~\ref{fig:viro_comp}, the pipeline consists of two main stages: (i) a pre-execution stage that translates the query $Q$ into a symbolic program $P$ (detailed in Section~\ref{sec:method:pre-exe}), and (ii) a program execution stage that runs this program $P$ on the image $I$ to localize the referent (detailed in Section~\ref{sec:method:exe}). The program $P$ is constructed using verification operators, with additional details on their functionality presented in the following section.

% A fundamental challenge is that program execution can fail at any step if a referent is absent or a specified relation does not hold, yet previous approaches often provide no insight into which specific reasoning step caused the failure.
% Our primary contribution addresses this challenge through the design of robust operators, enabling traceable execution, as detailed in the following section.

\iffalse
\begin{figure*}[t!]
     \centering
    
     \begin{subfigure}[b]{0.9\textwidth}
         \centering
         \includegraphics[width=\textwidth]{figures/figure2_a.pdf}
         \vspace{-1.5em}
         \caption{A CLIP-based uncertainty filter within \texttt{FIND} operator to eliminate FPs from OVD proposals.}\label{fig:viro}
     \end{subfigure}
     \vspace{1em}
     \begin{subfigure}[b]{0.9\textwidth}
         \centering
         \includegraphics[width=\textwidth]{figures/figure2_b.pdf}
         \caption{A pipeline comparison of pre-execution and execution stages.}\label{fig:pipeline}
     \end{subfigure}
    \vspace{-1em}
    \caption{Overview of \textsc{VIRO}. (a) Details of the \texttt{FIND} operator with a CLIP-based uncertainty filter. (b) The decoupled \textsc{VIRO} pipeline and comparison with compositional reasoning approaches.}
    \vspace{-1em}
\end{figure*}
\fi

\subsubsection{Verification Reasoning Operators (VROs)}\label{sec:method:vro}

We define a finite set of primitive operators, termed VROs and denoted by $\mathcal{O}$, which form the building blocks of our VIRO framework as summarized in Table~\ref{tab:operator}. A program is represented as $P = (o_1, o_2, \dots, o_T)$, where $o_t \in \mathcal{O}$ and $T$ denotes the number of program lines. Each operator in $\mathcal{O}$ is designed not only to perform a reasoning step but also to self-verify execution: if its verification condition is not satisfied, the operator returns an empty set ($\varnothing$), enabling early termination of the entire pipeline.
% If an operator determines that its verification condition is not satisfied, the operator returns an empty set ($\varnothing$), enabling early termination of the entire pipeline.

We categorize these operators into four functional categories:
(i) \textbf{Identification operators}, such as \texttt{FIND} and \texttt{PROPERTY}, which detect candidates and refine entities based on specified attributes;
(ii) \textbf{Absolute spatial operators}, such as \texttt{LOCATE}, \texttt{SIZE}, \texttt{ORDER}, and \texttt{ABSOLUTE\_DEPTH}, which reason about position and scale in absolute terms;
(iii) \textbf{Relative spatial operators}, such as \texttt{FIND\_DIRECTION}, \texttt{FIND\_NEAR}, \texttt{FIND\_INSIDE}, and \texttt{RELATIVE\_DEPTH}, which capture spatial relationships between multiple entities;
(iv) \textbf{Termination operator}, namely \texttt{RESULT}, which concludes the program by mapping the selected object into the answer space.
Further details of each predefined operator, including their arguments, are provided in Table~\ref{tab:operator}.
To illustrate how the verification module works, we detail two key examples below.

\vspace{-.5em}
\paragraph{Uncertainty Verification (UV) in \texttt{FIND} Operator.}
The \texttt{FIND} operator takes an argument \texttt{object\_name}, a noun phrase $l$ (\textit{e.g.}, ``guy'') in the query $Q$ (\textit{e.g.}, ``A middle guy in red'').
It invokes an OVD model $D$ (\textit{e.g.}, Grounding DINO~\citep{liu2024grounding}) on image $I$ with label $l$ to generate a (possibly empty) set of proposals, \textit{i.e.}, $\{B_j\}_{j=1}^M \leftarrow D(I, l)$.
While state-of-the-art OVDs offer powerful zero-shot grounding, they can yield high-confidence false positives (FPs) that are visually or semantically similar to $l$, which can propagate error through the reasoning pipeline, as shown in Figure~\ref{fig:viro_comp}.

To mitigate FP proposals, 
we integrate a lightweight CLIP-based verification module within the \texttt{FIND} operator. 
This module provides a secondary check on the OVD's proposals by leveraging CLIP's discriminative power of binary classification tasks, effectively filters uncertain outputs while adding minimal computational overhead.
Specifically, 
for each proposal $B_j$, we crop its corresponding image region, denoted as $I_j$.
We predefine a bank of $K$ common categories $\mathcal{C}=\{c_1, c_2, \dots, c_K\}$ that are well represented in CLIP and serve as negative anchors for verification, where $K$ denotes the number of negative anchors.
The verification score for $B_j$ is the average probability of being the target $l$ when compared one-on-one against each $c_k\in\mathcal{C}$:
\begin{align}\label{eq:score}
    S(l\,|\!\,I_j)\!=\!\frac{1}{K}\!\sum_{k=1}^{K}\!
    \frac{\exp(\operatorname{sim}(I_j,l)/\tau)}
    {\exp(\operatorname{sim}(I_j,l)/\tau)+\exp(\operatorname{sim}(I_j,c_k)/\tau)} \;,
\end{align}
where $\operatorname{sim}(\cdot,\cdot)$ denotes cosine similarity between CLIP image/text embeddings and $\tau>0$ is a temperature.
We accept $B_j$ as $l$ only if $S(l\mid I_j)\ge \delta_{l}$, where $\delta_{l}$ is a fixed or label-specific adaptive threshold. There is a trade-off in setting the threshold: when it is set close to $0.5$, true positives (TP) may be filtered out, while FPs may remain unfiltered. 
Since CLIP can exhibit inherent bias toward labels that were well-represented in its training data, this can affect the accuracy of the thresholding process. To mitigate this, we use ImageNet~\citep{deng2009imagenet} as an auxiliary dataset for per-label threshold calibration.  
Details of the calibration process are provided in Appendix~\ref{appendix:threshold}. To summarize, $S(l|I_j)$ serves as an uncertainty filter, assessing the degree of alignment between the given label $l$ and the regions $I_j$ proposed by OVD, and filtering out uncertain proposals.

\vspace{-.5em}
\paragraph{Logical Verification (LV) in \texttt{FIND\_DIRECTION} Operator.}

The \texttt{FIND\_DIRECTION} operator takes three arguments: \texttt{object}, \texttt{reference\_object}, and \texttt{direction}. \texttt{object} refers to the target object that we are trying to find, while the \texttt{reference\_object} is the object used as a reference for comparison.
It performs a geometric test over all input candidates to verify whether each \texttt{object} proposal satisfies the specified spatial relation with respect to at least one \texttt{reference\_object}.

\subsubsection{Pre-execution Stage: Program Generation}\label{sec:method:pre-exe}
% \begin{figure}[!t]
%   \centering
% \includegraphics[width=0.95\linewidth,trim=0 0 0 0,clip]{figures/VIRO_overview.pdf}
%   \caption{}
%   \label{fig:pipeline}
% \end{figure}

The pre-execution stage translates a given natural-language query $Q$ into a machine-executable symbolic program $P$, which is composed of our primitive operators defined in 
% Section~\ref{sec:method:exe:operator}. 
Section~\ref{sec:method:vro}.
This translation into a structured format is crucial for ensuring robust execution, as the inherent complexity of natural language makes direct machine interpretation unreliable. 
This process is accomplished through two key components: a program generator that produces an initial program, and a program validator that subsequently ensures its syntactic correctness.

\vspace{-.5em}
\paragraph{Program Generation.}
We leverage a Large Language Model (LLM) to translate the natural language query $Q$ into a symbolic program $P$. We guide this process using a few-shot prompting strategy as in \cite{gupta2023visual}, where the prompt contains a set of exemplars $m$ demonstrating the desired query-to-program mapping:
% . The generated program $P$ consists of a linear sequence of operators $(o_1, o_2, ..., o_T)$, where each operator is drawn from our predefined set of Verification-Integrated Operators $\mathcal{O}$, detailed in Section~\ref{sec:method:operator}.
% Formally, this generation is expressed as follows:
\begin{align}
    P\!=\!\text{LLM}(Q|m) = (o_1, o_2, \ldots, o_T), \; \text{where } o_t \in \mathcal{O} \;.
\end{align}

\vspace{-.5em}

\paragraph{Program Validation.}

LLMs can occasionally produce programs with syntactic errors such as malformed syntax, wrong operator names, or mismatched arguments. To mitigate this, we introduce a program validator
that ensures strict conformance to our predefined operator-based grammar. Unlike ViperGPT's open-ended Python code, which is prone to runtime or library-call errors, our approach restricts the output space to a symbolic structure.

If validation fails, a concise diagnostic feedback is provided to the LLM to trigger an immediate self-revision. This structured validation loop, combined with our compact operator set, allows our symbolic approach to achieve significantly lower failure rates compared to existing methods, as evidenced in Section~\ref{sec:exp:efficiency}. Detailed diagnostic rules are provided in Appendix~\ref{appendix:validator}.

\iffalse
While powerful, LLMs can occasionally produce programs with syntactic errors such as malformed syntax, wrong operator names, or mismatched arguments. Unlike ViperGPT~\citep{suris2023vipergpt}, which relies on compiling Python code as shown in Figure~\ref{fig:viro_comp}, our validator simply checks for conformance to our predefined operator-based grammar. If validation fails, a concise diagnostic is fed back to the LLM in a new prompt to trigger a revision. See Appendix~\ref{appendix:validator} for details.
Furthermore, symbolic approaches achieve low failure rates, as shown in Section~\ref{sec:exp:efficiency}.
\fi

\subsubsection{Execution Stage: Program Interpretation}\label{sec:method:exe}

The program interpreter executes the symbolic program $P = (o_1, o_2, \dots, o_T)$ sequentially, invoking the corresponding operator at each step. Each operator relies on a built-in model to function properly, and we leverage the following models: GroundingDINO~\citep{liu2024grounding} or GLIP~\citep{li2022grounded}, CLIP~\citep{radford2021learning}, and DepthAnything~\citep{yang2024depth}.
Execution continues until either (i) all operators are applied and \texttt{RESULT} maps the final candidates to an answer box, or (ii) some operator returns $\varnothing$, yielding a no-target outcome for the current image and immediately terminating execution.
The latter corresponds to an early-exit, which occurs when built-in verification rejects all proposals (\textit{e.g.}, \texttt{FIND} identifies there is no valid object; \texttt{FIND\_DIRECTION} finds no relation). 
This mechanism enables explicit no-target handling and reduces unnecessary computation, as demonstrated in Section~\ref{sec:exp:ablation}, thereby highlighting the robustness of our verification-integrated design.

\subsubsection{Decoupled Neuro-Symbolic Approach}\label{sec:method:benefit}
As shown in Figure~\ref{fig:viro_comp}, \textsc{VIRO} adopts a decoupled design that separates program generation from execution. In contrast, methods such as HYDRA~\citep{ke2024hydra} and NAVER~\citep{cai2025naver} entangle program synthesis for a query $Q$ with image execution. Consequently, even when the query is identical across $N$ images, these systems regenerate a reasoning program for each image $I_i$, incurring $N$ separate synthesis operations. VIRO generates the program once and reuses it across all images, enabling low-latency operation. Empirical results are reported in Section~\ref{sec:exp:efficiency}.

% \iffalse
% \subsubsection{Program Interpreter}
% while also improving \textbf{efficiency} by terminating execution early when no valid candidates remain.

% This verification-aware execution enables the system to maintain both robustness with  by abstaining in no cases) and \textbf{efficiency} (by terminating early when no valid candidates remain).

% \paragraph{No-target.} Formally, given an operator $o_t$ with inputs $x_{t-1}$ and verification function $v_t$, the interpreter computes

% Execution continues until either (i) all operators are successfully applied, in which case the \texttt{RESULT} operator maps the final proposals to the answer space \hj{max}, or (ii) early-exit is triggered.

% \paragraph{Early-Exit.} 
% If at any point the verification module rejects all candidates (e.g., no bounding box passes filtering in \texttt{FIND}, or no spatial relation holds in \texttt{FIND\_DIRECTION}), the interpreter triggers an early-exit, returning $\varnothing$ to indicate that the query cannot be grounded in the current image. This mechanism allows the system to gracefully handle \emph{no-target scenarios} while avoiding unnecessary computation, as you can see the effect in Section~\ref{sec:exp:ablation}.
% \fi

\section{Experiment}\label{sec:exp}

\begin{table*}[htb!]
\centering
\caption{
Comparison of REC methods on no-target robustness using gRefCOCO no-target and standard REC performance using RefCOCO across the TestA and TestB splits. 
For each split, we report \textbf{Balanced Accuracy} (Bal. Acc.), along with its component metrics \textbf{True Negative Rate} (TNR; N-acc), and \textbf{True Positive Rate} (TPR; Acc@0.5).
}
% \caption{
  % Comparison of REC methods on robustness using the gRefCOCO no-target testA split, and on standard REC performance using the RefCOCO testA split. %TNR, FPR, and TPR denote True Negative Rate, False Positive Rate, and True Positive Rate respectively. % The best performance is in bold, and the second-best is \underline{underlined}.
% }
\vspace{-.5em}
\sisetup{table-format=2.2} 
\begin{adjustbox}{width=0.81\linewidth}
\begin{tabular}{lcccccc}
\toprule
 & \multicolumn{3}{c}{TestA} & \multicolumn{3}{c}{TestB} \\
 \cmidrule(lr){2-4} \cmidrule(lr){5-7} 
 % & & gRefCOCO & RefCOCO & & gRefCOCO & RefCOCO  \\
% \cmidrule(lr){3-3} \cmidrule(lr){4-4} \cmidrule(lr){6-6} \cmidrule(lr){7-7} 
% Method & Bal. Acc. $\uparrow$ & TNR $\uparrow$ & TPR  $\uparrow$ & Bal. Acc. $\uparrow$ & TNR  $\uparrow$ & TPR  $\uparrow$ \\
Method & Bal. Acc. $\uparrow$ & TNR {\scriptsize (gRef)} $\uparrow$ & TPR {\scriptsize (Ref)} $\uparrow$ & Bal. Acc. $\uparrow$ & TNR {\scriptsize (gRef)} $\uparrow$ & TPR {\scriptsize (Ref)} $\uparrow$ \\
\midrule

% --- Header --- % OK
\multicolumn{7}{l}{\textit{\textbf{Fully Supervised REC}}~\citep{he2023grec}} \\ 
Qwen2.5-VL-72B-AWQ$^{\dagger}$~\citep{bai2025qwen2} & 69.5 & 47.3 & 91.7 & 66.8 & 45.1 & 88.4 \\
GREC-MDETR-R101~\citep{kamath2021mdetr} & 62.0 & 34.5 & 89.6 & 56.2 & 31.0 & 81.4 \\
GREC-UNINEXT-R50~\citep{yan2023universal} & 70.4 & 49.3 & 91.5 & 67.6 & 48.2 & 86.9 \\
\midrule
% --- Second header ---
\multicolumn{7}{l}{\textit{\textbf{Proposal-based REC}}} \\ %$^\dagger$}}} \\
ReCLIP~\citep{subramanian2022reclip}& 23.5 & 0.0 & 47.0 & 22.6 & 0.0 & 45.2 \\
SS-CLIP~\citep{han2024zero} & 33.3 & 0.0 & 66.5  & 27.5 & 0.0 & 54.9 \\
% SS-FLAVA~\citep{han2024zero} & - & - & - & - \\
GroundVLP~\citep{shen2024groundvlp} & 30.7 & 0.0 & 61.3  & 21.8 & 0.0 & 43.5  \\
% --- Third header ---
\midrule
\multicolumn{7}{l}{\textit{\textbf{Detector-based REC}}} \\
GLIP-L~\citep{li2022grounded} & 37.2 & 21.7 & 52.6 & 30.0 & 18.2 & 41.8 \\
GroundingDINO-T~\citep{liu2024grounding} & 40.0 & 22.8 & 57.2 & 29.6 & 16.0 & 43.2 \\
\midrule
% --- header ---
\multicolumn{7}{l}
{\textit{\textbf{Compositional Reasoning REC with GroundingDINO}}} \\
ViperGPT~\citep{suris2023vipergpt} & 33.4 & 0.2 & 66.7 & 27.4 & 0.1 & 54.6 \\
HYDRA~\citep{ke2024hydra} & 35.2 & 7.5 & 62.8 & 34.7 & 7.0 & \textbf{62.4} \\
NAVER~\citep{cai2025naver} & 33.8 & 3.4 & 64.2 & 30.0 & 1.8 & 58.2 \\
\rowcolor[HTML]{EFEFEF} \textbf{VIRO (Ours)} & \textbf{61.1} & \textbf{50.2} & \textbf{71.9} & \textbf{56.9} & \textbf{52.9} & 60.8 \\
% \midrule
% \multicolumn{5}{p{1.1\textwidth}}{\footnotesize $^{\dagger}$ Most baselines adopt as \textit{forced-choice} design, which makes handling \textit{no-target} cases particularly challenging. We provide per-algorithm heuristic analyses in Appendix~\ref{appendix:heuristic_notarget}.} \\
\bottomrule
\multicolumn{7}{p{.95\linewidth}}{\footnotesize \textit{Note:} $^{\dagger}$We append a negative instruction, \textit{e.g.,} ``If there is no object, return []''. Without this prompt, the model struggles to detect no-target cases, reducing TNR 3.1\% (TPR 94.7\%) on TestA.} \\
\end{tabular}
\end{adjustbox}
\label{tab:robustness}
\vspace{-1.0em}
\end{table*}

In this section, we present a comprehensive evaluation of our VIRO pipeline. The details of the experimental setup are described in Section~\ref{sec:exp:setup}. 
We then present our main results in terms of robustness, efficiency, and scalability in Section~\ref{sec:exp:main}. Following this, we conduct extensive ablation studies in Section~\ref{sec:exp:ablation}. %, analyzing the contribution of each component.%, and provide qualitative examples in Section~\ref{sec:exp:qualitative}.

\subsection{Experimental Setup}\label{sec:exp:setup}
\paragraph{Implementation Details.}
We primarily follow the official implementations of each baseline, using their default hyperparameter settings. 
Unless otherwise noted, detection thresholds for open-vocabulary detectors are fixed based on validation performance on RefCOCO.
% : $0.2$ for GroundingDINO-T and $0.5$ for GLIP-L. 
For all program generation, we use Qwen2.5-72B-Instruct-AWQ~\citep{qwen2.5}, known for strong code-generation, to ensure a fair comparison with Python-code baselines such as ViperGPT. 
% \add{Additional results with Llama3.1-8B-Instruct~\citep{grattafiori2024llama3} are reported in Appendix~\ref{appendix:llm}.}

\paragraph{Dataset and Evaluation Metrics.} 
We evaluate our framework on both no-target scenarios and standard benchmarks.
For the no-target setting, we use the gRefCOCO dataset~\citep{he2023grec} no-target split, which contains referring expressions that do not correspond to any object in the image. This setting allows us to directly evaluate the model’s ability to suppress incorrect predictions in the absence of a valid target.
For standard REC, we evaluate our method on widely used REC benchmarks, including RefCOCO/+~\citep{yu2016modeling}, and RefCOCOg~\citep{mao2016generation}.
We further assess the consistency of linguistic structure comprehension on RefAdv~\citep{akula2020words}, and extend evaluation to the video domain with RefEgo~\citep{kurita2023refego}, which includes both target-present and no-target cases.
We further show that VIRO can be modularly extended to GQA~\cite{hudson2019gqa}, suggesting potential applicability beyond REC, with detailed results provided in Appendix~\ref{appendix:gqa}.
% We further evaluate VIRO on GQA~\cite{hudson2019gqa} to examine its ability to generalize beyond REC to compositional visual reasoning, with detailed results provided in Appendix~\ref{appendix:gqa}.
% \hj{
% We also evaluate VIRO on GQA to assess compositional visual reasoning beyond REC. Detailed task formulation and evaluation protocol are provided in Appendix~\ref{appendix:gqa}.}
% We additionally include GQA as an auxiliary benchmark to evaluate compositional visual reasoning beyond referring expression comprehension; details are deferred to Appendix A.3.}

We evaluate both no-target robustness and standard REC accuracy via \textbf{Balanced Accuracy}, defined as $(\text{TPR} + \text{TNR})/2$. 
It provides a holistic measure by equally weighing the ability to localize existing targets and reject non-existing ones.
The component metrics are defined as follows:
\begin{itemize}[leftmargin=10pt, itemsep=0pt, topsep=0pt]
    \item \textbf{True Positive Rate (TPR)} = $\tfrac{TP}{TP+FN}$, measuring accuracy on target-present samples. 
    This is equivalent to the standard Acc@0.5, used in standard REC, where a prediction is correct if IoU between the predicted and ground-truth bounding boxes exceeds $0.5$.  
    \item \textbf{True Negative Rate (TNR)} = $\tfrac{TN}{TN+FP}$, measuring accuracy on target-absent samples, often referred to as no-target accuracy (N-acc).  
    % \item \textbf{False Positive Rate (FPR)} = $\tfrac{FP}{TN+FP} = 1 - \text{TNR}$, quantifying how often the model incorrectly predicts a target in images where none exist.  
\end{itemize}

% forcing the model to output a candidate 
\vspace{-1em}
\paragraph{Baselines.}
We group existing REC approaches into four categories, according to their underlying assumptions. 
\textbf{Fully supervised REC} includes methods trained with full annotations on RefCOCO/+/g, while GREC~\citep{he2023grec} additionally uses no-target supervision to handle target-absent cases.
% \textbf{Fully supervised REC} includes methods trained end-to-end with full annotations on RefCOCO/+/g. % as well as on the gRefCOCO no-target dataset for handling no-target cases.
\textbf{Proposal-based REC} methods first parse the referring expression to extract key linguistic components and then align them with candidate region proposals.
Such approaches inherently force the model to select one of the proposals, which makes handling no-target cases intrinsically difficult. 
\textbf{Detector-based REC} leverages large-scale pretrained grounding detectors to directly match textual phrases with image regions in an end-to-end manner, without explicit proposal ranking. 
For this category, we select GroundingDINO-T~\citep{liu2024grounding} and GLIP-L~\citep{li2022grounded} as representative methods. 
Importantly, neither model was trained on the MSCOCO captions underlying our evaluation benchmarks; accordingly, we evaluate them in a zero-shot setting to ensure a fair comparison.
% A key criterion for their selection is that neither model was trained on the MSCOCO captions that are the source of our evaluation benchmarks, thereby ensuring a fair comparison.
Finally, \textbf{compositional reasoning REC}, which serves as our primary point of comparison, explicitly parses and executes the linguistic structure to localize referents through multi-step reasoning. More explanations of the baselines and additional details of the experimental setup are provided in Appendix~\ref{appendix:exp}.

\subsection{Main Results}\label{sec:exp:main}
We evaluate our framework along three key dimensions: 
(i) robustness in handling \textit{no-target} cases (Section~\ref{sec:exp:robustness}), 
(ii) efficiency in terms of failure rate and execution latency (Section~\ref{sec:exp:efficiency}), and 
(iii) scalability, highlighting the benefits of our decoupled pipeline (Section~\ref{sec:exp:scalability}).
Furthermore, we extend our evaluation to a real-world egocentric setting with the RefEgo dataset, which includes both target-present and no-target cases (Section~\ref{sec:exp:refego}).

\subsubsection{Robustness on No-Target Cases}\label{sec:exp:robustness}
Table~\ref{tab:robustness} reports results on the RefCOCO TestA and TestB splits, where TestA contains person referents and TestB contains object-centric referents.
The proposal-based baselines yield near-zero TNR because they operate as forced-prediction systems, selecting one region from a pre-generated proposal pool (e.g., Faster R-CNN in MAttNet~\citep{ren2016faster,yu2018mattnet}) whenever candidate boxes are available.
Similarly, lacking abstention mechanisms, detector-based and compositional REC methods often select hallucinated OVD detections.
This forced-prediction setup introduces an implicit trade-off: under target-present assumptions, such methods may appear stronger than in no-target-aware settings.
We revisit this point in Appendix~\ref{appendix:forced_choice}, showing that our method also attains higher TPR when abstention is disallowed.
% This forced prediction creates an implicit trade-off: it severely degrades no-target robustness while artificially inflating TPR on standard benchmarks by rewarding guesses. 
% We validate this inflation in Appendix~\ref{appendix:forced_choice}, showing our method also exhibits inflated TPRs under a forced-prediction setting.
Since baseline models inherently lack no-target handling, the criteria we established to evaluate their no-target cases are detailed in Appendix~\ref{appendix:explanation_notarget}.

In contrast, VIRO achieves 61.1\% and 56.9\% Balanced Accuracy on TestA and TestB, respectively, substantially outperforming all compositional reasoning baselines without REC fine-tuning. These gains stem from verification-integrated operators that enable abstention when no valid referent is present. Crucially, VIRO remains competitive with fully supervised methods (e.g., GREC-UNINEXT) despite not using no-target annotations, demonstrating robust zero-shot visual grounding. Results on the remaining splits are provided in Appendix~\ref{appendix:all_results}.

% These gains stem from verification-integrated operators that allow the model to abstain when no valid referent is present, rather than forcing a prediction. Importantly, VIRO remains competitive with fully supervised methods despite not relying on REC fine-tuning or no-target annotations, demonstrating robust zero-shot visual grounding across both person-centric and object-centric settings.

\subsubsection{Efficiency of Compositional Reasoning on Standard REC Benchmarks}\label{sec:exp:efficiency}
\begin{table*}[t!]
\centering
\caption{
Comparison of compositional reasoning accuracy and efficiency on REC benchmarks.
All results are evaluated using Qwen2.5-72B-Instruct-AWQ on RefCOCO/RefCOCO+ (testA) and RefCOCOg (test).
TPR (Acc@0.5) is measured both including (Inc.$\uparrow$) and excluding (Exc.$\uparrow$) failure cases and FR denotes the Failure Rate (\%).
Runtime (query/s) is measured on the RefCOCO testA split using an NVIDIA RTX A6000 GPU; E2E refers to total per-query latency, while Exec. refers to the execution-stage only.
}
\vspace{-.3em}
\begin{adjustbox}{width=.89\linewidth}
\begin{tabular}{lcccccccccccc}
\toprule
 & \multicolumn{3}{c}{RefCOCO} & \multicolumn{3}{c}{RefCOCO+} & \multicolumn{3}{c}{RefCOCOg} & \\
 \cmidrule(lr){2-4} \cmidrule(lr){5-7} \cmidrule(lr){8-10} 
Method & FR $\downarrow$ & Exc.$\uparrow$ & Inc.$\uparrow$ & FR $\downarrow$ & Exc.$\uparrow$ & Inc.$\uparrow$ & FR $\downarrow$ & Exc.$\uparrow$ & Inc.$\uparrow$ & 
E2E $\downarrow$  & Exec. $\downarrow$  \\
\midrule
\multicolumn{5}{l}{\textbf{\textit{Fully Supervised REC}}} \\
Qwen2.5-VL-72B-AWQ~\citep{bai2025qwen2} & 0.12 & 94.4 & 94.3 & 0.17 & 91.8 & 91.6 & 0.11 & 89.1 & 89.0 & 9.39 & 9.39\\
\midrule
\multicolumn{5}{l}{\textbf{\textit{Detector-based REC}}} \\
GLIP-L~\citep{li2022grounded} & 0.00 & 52.6 & 52.6 & 0.00 & 48.6 & 48.6 & 0.00 & 52.6 & 52.6 & 0.81 & 0.81 \\
GroundingDINO-T~\citep{liu2024grounding} & 0.00 & 57.2 & 57.2 & 0.00 & 57.6 & 57.6 & 0.00 & 59.5 & 59.5 & 0.20 & 0.20 \\
% \midrule
% \multicolumn{5}{l}{\textbf{\textit{Compositional Reasoning REC with GLIP}}} \\
% ViperGPT~\citep{suris2023vipergpt} & 3.32 & 72.0 & 69.6 & 3.28 & 65.7 & 61.3 & 6.31 & 69.6 & 65.2 & \hj{} & 1.47 \\
% HYDRA~\citep{ke2024hydra} & 17.27 & 73.1 & 60.5 & 17.06 & 60.6 & 48.7 & 21.44 & 67.6 & 53.1 & 24.18 & 20.00 \\
% NAVER~\citep{cai2025naver} & 14.85 & 73.4 & 62.5 &14.67 & 62.7 & 50.7 & 25.90 & 70.0 & 51.9 & \hj{} & 5.88 \\
% \cellcolor[HTML]{EFEFEF}\textbf{VIRO (Ours)} & \cellcolor[HTML]{EFEFEF}0.07 & \cellcolor[HTML]{EFEFEF}72.8 & \cellcolor[HTML]{EFEFEF}\textbf{72.8} & \cellcolor[HTML]{EFEFEF}0.07 & \cellcolor[HTML]{EFEFEF}63.7 & \cellcolor[HTML]{EFEFEF}\textbf{63.7}& \cellcolor[HTML]{EFEFEF}0.30 & \cellcolor[HTML]{EFEFEF}67.0 & \cellcolor[HTML]{EFEFEF}\textbf{66.8} & \cellcolor[HTML]{EFEFEF}\hj{} & 
% \cellcolor[HTML]{EFEFEF}0.80\\
\midrule
\multicolumn{5}{l}{\textbf{\textit{Compositional Reasoning REC with GroundingDINO}}} \\
ViperGPT~\citep{suris2023vipergpt} & 3.45 & 66.7 & 64.4 & 6.83 & 61.7 & 57.5 & 6.03 & 65.7 & 61.7 & 30.19 & 1.49 \\
HYDRA~\citep{ke2024hydra} & 28.46 & 62.8 & 44.9 & 35.96 & 58.4 & 37.4 & 32.37 & 67.1 & 45.4 & 37.06 & 30.57 \\
NAVER~\citep{cai2025naver} & 6.01 & 64.2 & 60.3 & 7.39 & 60.1 & 55.6 & 9.74 & 68.4 & 55.8 & 7.74 & 7.08 \\
\cellcolor[HTML]{EFEFEF}\textbf{VIRO (Ours)} & \cellcolor[HTML]{EFEFEF}0.07 & \cellcolor[HTML]{EFEFEF}71.9 & \cellcolor[HTML]{EFEFEF}\textbf{71.9} & \cellcolor[HTML]{EFEFEF}0.00 & \cellcolor[HTML]{EFEFEF}63.3 & \cellcolor[HTML]{EFEFEF}\textbf{63.3} & \cellcolor[HTML]{EFEFEF}0.30 & \cellcolor[HTML]{EFEFEF}66.6 & \cellcolor[HTML]{EFEFEF}\textbf{66.3} & \cellcolor[HTML]{EFEFEF}12.92 & 
\cellcolor[HTML]{EFEFEF}0.71 \\
\bottomrule
\end{tabular}
\end{adjustbox}
\label{tab:efficiency}
\vspace{-1em}
\end{table*}

Table~\ref{tab:efficiency} compares VIRO with compositional reasoning REC baselines on standard REC benchmarks in terms of accuracy, program failure rate (FR), and runtime efficiency. Due to space constraints, RefAdv results for linguistic understanding are reported in Appendix~\ref{appendix:refadv}.
To rigorously assess reliability, the inclusive accuracy (Inc.) strictly penalizes program failures as incorrect predictions~\cite{cai2025naver}.
VIRO exhibits a remarkably low program failure rate of less than 0.3\%, making its Inc. and Exc. accuracies nearly identical.
In contrast, HYDRA and NAVER frequently fail due to syntactically invalid LLM-generated programs or execution timeouts when no answer is found within their default iteration budgets (7 and 5, respectively). VIRO's integrated validator, with a maximum of 5 iterations, ensures that nearly all generated programs are executable, overcoming a critical bottleneck of previous compositional approaches. 

Moreover, VIRO incurs lower execution runtime (Exec.), indicating that its structured operator-based reasoning introduces little additional computational overhead. This makes VIRO well suited for large-scale image processing per query, as analyzed in the following section.
Details regarding pre-execution acceleration via API-based LLMs (\textit{e.g.}, GPT-4o/mini~\cite{hurst2024gpt}) and baseline VRAM requirements are provided in Appendices~\ref{appendix:runtime}~and~\ref{appendix:models}, respectively.
% We additionally show that the pre-execution time can be further accelerated by replacing the local Qwen2.5-72B-Instruct-AWQ with faster LLM models such as GPT-4o in Appendix~\ref{appendix:runtime}. Details regarding the underlying visual-language components used in the compositional baselines and their VRAM requirements are provided in Appendix~\ref{appendix:models}.

% Furthermore, VIRO achieves high throughput (FPS) during the execution stage than other compositional reasoning methods. This demonstrates that VIRO incorporates a sophisticated reasoning layer via its operators without imposing a heavy computational burden, making it highly suitable for processing a large number of images as further analyzed in the following section. Additional analysis of pre-execution and execution time is presented in Appendix~\ref{appendix:runtime}, along with
% the models used in the compositional baselines and their
% VRAM requirements in Appendix~\ref{appendix:models}.

% \hj{models - if possible compare}

\subsubsection{Scalability in 1-Query--\texorpdfstring{$N$}{N}-Images}\label{sec:exp:scalability}

\begin{figure}[!tb]
    \centering
    \includegraphics[width=0.43\textwidth]{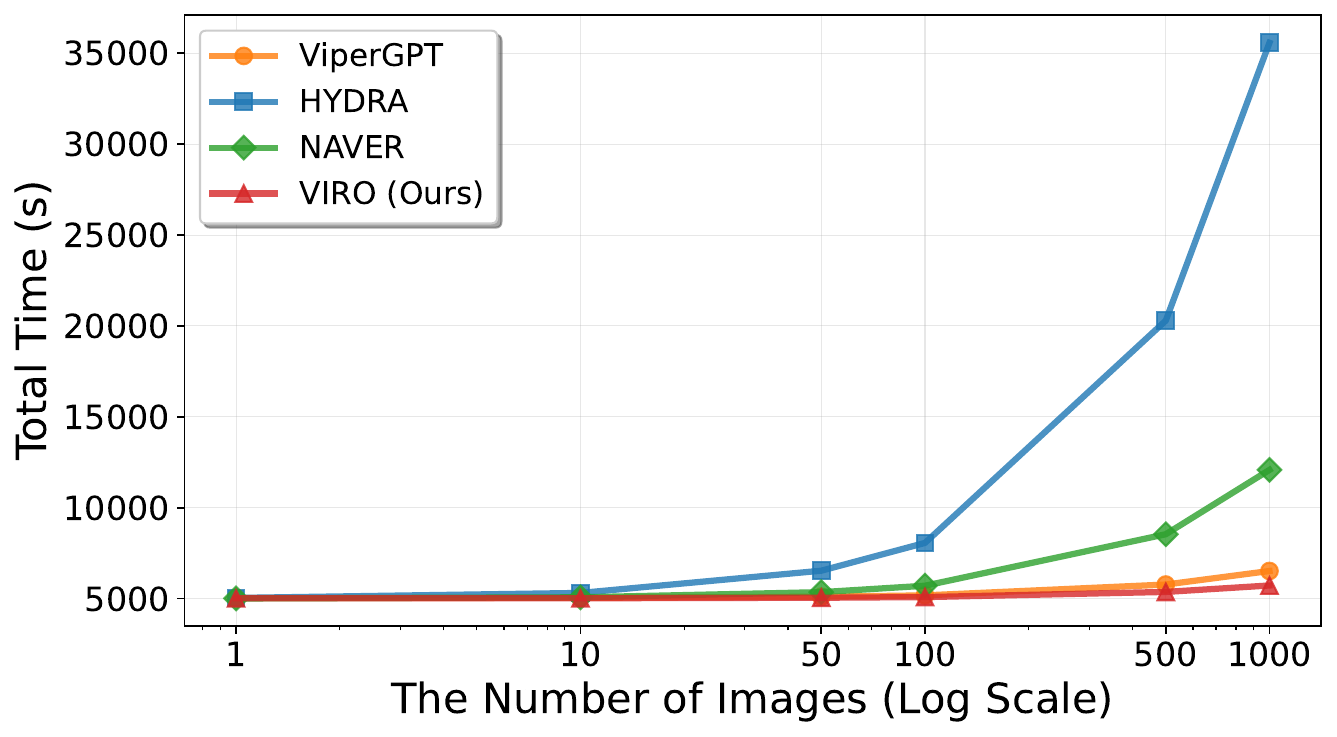}
    \vspace{-.6em}
    \caption{Total inference time (s) in a 1-query-$N$-images setting, with the $x$-axis on a logarithmic scale.}
    \label{fig:scalability}
    \vspace{-1.3em}
\end{figure}

% \begin{wrapfigure}{r}{0.44\textwidth} % r: right align
%     \centering
%     \vspace{-1.2em}
%     \hspace{-1em}
%     \includegraphics[width=\linewidth]{figures/ablation_scale.pdf} 
%     \vspace{-.5em}
%     \caption{Total time in a 1-query-$N$-images setting, with the $x$-axis on a log scale. }
%     \label{fig:scalability}
%     \vspace{-1em}
% \end{wrapfigure}

We evaluate the scalability of VIRO in a 1-query-$N$-images setting---a crucial scenario for real-world applications like robotic visual search.
%, where a single query is used to localize a target across multiple $N$ images. This scenario is crucial for real-world applications such as robotic visual search.
As shown in Figure~\ref{fig:scalability}, VIRO and ViperGPT demonstrate exceptional scalability. Due to its decoupled architecture, the program generation is performed only once per query, \textit{i.e.}, $T_{\text{total}}=T_{\text{pre-execution}}+N\times T_{\text{execution}}$.
In contrast, entangled methods like HYDRA and NAVER force program regeneration for each image, \textit{i.e.}, $T_{\text{total}}=N\times T_{\text{pre-execution}}+N\times T_{\text{execution}}$.
This severe computational overhead makes them impractical for large-scale tasks, highlighting the critical advantage of our decoupled design for low-latency visual reasoning.
% methods like HYDRA and NAVER entangle these stages, forcing program regeneration for each image. 
% Their total time is calculated as $T_{\text{total}}=N\times T_{\text{pre-execution}}+N\times T_{\text{execution}}$, leading to a non-linear growth in processing time that is impractical for large-scale tasks.
% This result highlights the critical advantage of our decoupled design for achieving low-latency visual reasoning at scale.

\subsubsection{Robustness Evaluation on RefEgo}\label{sec:exp:refego}

\begin{table}[tb!]
\centering
\caption{Comparison of video-based REC performance on the RefEgo test set. Refer to Appendix~\ref{appendix:refego_val} for the corresponding validation set results.}%Quantitative results on the RefEgo dataset. Accuracy (\%) is evaluated under two conditions: including all predictions (Inc.$\uparrow$) and excluding model failure cases (Exc.$\uparrow$). FR denotes the overall Failure Rate (\%).}
\vspace{-.5em}
\begin{adjustbox}{width=1.0\linewidth}
\begin{tabular}{lccc}
\toprule
& \multicolumn{2}{c}{All Frames} & Target-present \\%multicolumn{2}{c}{target-present}  \\
\cmidrule(lr){2-3}\cmidrule(lr){4-4}
Method & mSTIoU & Acc@0.5+n & TPR (Acc@0.5) \\
\midrule
\multicolumn{4}{l}{\textbf{\textit{Fully Supervised with RefEgo}}} \\
MDETR+BH~\cite{kurita2023refego} & 36.9 & 51.1 & 53.0 \\
\midrule
\multicolumn{4}{l}{\textbf{\textit{Off-the-shelf RefCOCOg model}}} \\
OFA~\cite{wang2022ofa} & 15.4 & 29.3 & 27.1\\
MDETR~\cite{kamath2021mdetr} & 15.4 & 26.4 & 22.8 \\
\midrule
\multicolumn{4}{l}{\textbf{\textit{Compositional Reasoning REC with GroundingDINO}}} \\
ViperGPT~\cite{suris2023vipergpt} & 13.0 & 23.0 & 27.6 \\
\cellcolor[HTML]{EFEFEF}\textbf{VIRO (Ours)} & \cellcolor[HTML]{EFEFEF}\textbf{22.8} & \cellcolor[HTML]{EFEFEF}\textbf{51.9 }& \cellcolor[HTML]{EFEFEF}\textbf{36.2} \\
% MDETR+BH & 36.9 & 45.7 & 51.1 & 45.7 & 53.0 \\
% % \multicolumn{4}{l}{\textbf{\textit{Detector-based REC}}} \\
% OFA & 15.4 & 28.9 & 29.3 & 27.8 & 27.1\\
% MDETR & 15.4 & 25.6 & 26.4 & 22.9 & 22.8 \\
% \midrule
% \multicolumn{4}{l}{\textbf{\textit{Compositional Reasoning REC with GroundingDINO}}} \\
% ViperGPT & - & 21.6 & 23.0 & 25.7 & 27.6 \\
% VIRO (Ours) & 22.8 & 51.0 & 51.9 & 35.0 & 36.2 \\
% \cellcolor[HTML]{EFEFEF}\textbf{VIRO (Ours)} &
% \cellcolor[HTML]{EFEFEF}0.24 & \cellcolor[HTML]{EFEFEF}66.2 & \cellcolor[HTML]{EFEFEF}66.0  \\
\bottomrule
\end{tabular}
\end{adjustbox}
\label{tab:refego}
\vspace{-1.0em}
\end{table}

\begin{figure*}[t]
\centering
% ----- Left: Table -----
\begin{minipage}{0.63\linewidth}
\centering
\captionof{table}{
    Ablation study of the proposed verification components in VIRO on the gRefCOCO no-target and RefCOCO testA dataset. 
    `Fixed' refers to a fixed threshold, while `adaptive' refers to an adaptive threshold.
}
\vspace{-.5em}
\sisetup{table-format=2.2}
\begin{adjustbox}{width=\linewidth}
\begin{tabular}{lccc}
\toprule
\multicolumn{2}{c}{} & No Target Robustness & Standard REC  \\
\cmidrule(lr){3-3} \cmidrule(lr){4-4} 
Method & Balanced Acc. $\uparrow$ & TNR (N-acc) $\uparrow$ & TPR (Acc@0.5) $\uparrow$ \\
\midrule
Detector-only & 40.0 & 22.8 & 57.1 \\
+ Operators & 56.8 &  38.9 & \textbf{74.6} \\
+ LV & 57.0 &  39.3 & 74.6 \\
+ UV, fixed & 58.8  & 43.1 & 74.4 \\
\cellcolor[HTML]{EFEFEF}+ UV, adaptive
  & \cellcolor[HTML]{EFEFEF}\textbf{61.1} 
  & \cellcolor[HTML]{EFEFEF}\textbf{50.2} 
  & \cellcolor[HTML]{EFEFEF}71.9 \\
\bottomrule
\end{tabular}
\end{adjustbox}
\label{tab:ablation_verification}
\end{minipage}
\hfill
% ----- Right: Figure -----
\begin{minipage}{0.32\linewidth}
\centering
\includegraphics[width=\linewidth, trim=0 2 0 0, clip]{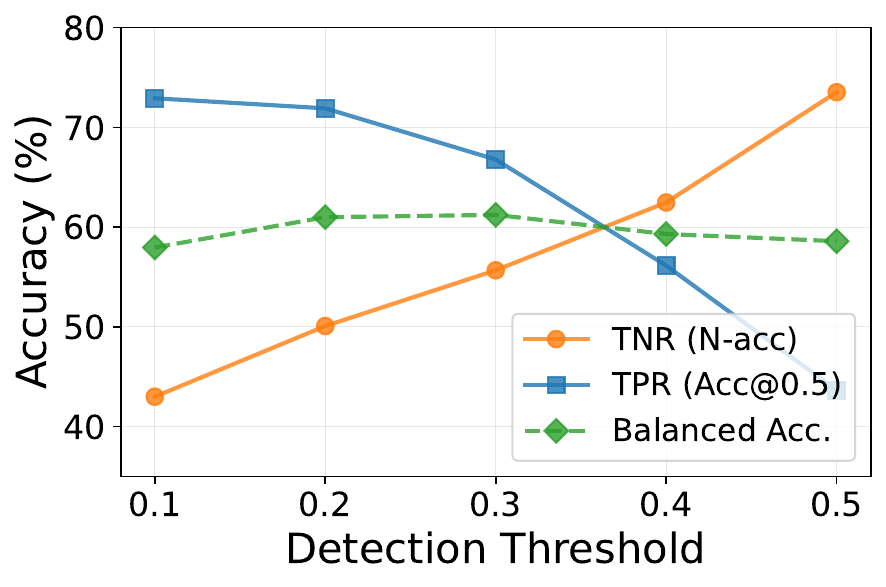}
\vspace{-2em}
    \caption{Analysis of the OVD detection threshold, illustrating the trade-off between TPR and TNR.}
    \label{fig:thres:detection}
\end{minipage}
\vspace{-1.5em}
\end{figure*}

RefEgo~\citep{kurita2023refego} is a video-based REC benchmark built upon Ego4D~\citep{grauman2022ego4d}, designed to evaluate grounding under egocentric motion and varying target visibility. Unlike static-image REC datasets, each video clip contains both target-present and no-target frames, requiring robust absent-target detection.
%a model to robustly determine not only where the target is, but also when it is absent.
Since this setting naturally corresponds to the $1$-query–$N$-images scenario (Section~\ref{sec:exp:scalability}), we restrict our comparison to methods whose pre-execution is decoupled from execution.
Following RefEgo, we report mean Spatio-Temporal IoU (mSTIoU) and ACC@0.5+n.  
For a video clip with $\mathcal{N}$ frames, STIoU is calculated as: % as the ratio between the sum of per-frame intersection areas and the sum of per-frame union areas: %$\mathrm{STIoU} = \sum_{i=1}^{\mathcal{N}} |p_i \cap t_i| / \sum_{i=1}^{\mathcal{N}} |p_i \cup t_i|,$
\begin{equation}
    \mathrm{STIoU} = 
    \frac{\sum_{i=1}^{\mathcal{N}} |p_i \cap t_i|}
         {\sum_{i=1}^{\mathcal{N}} |p_i \cup t_i|} \;,
\end{equation}
where $p_i$ and $t_i$ denote the predicted and ground-truth regions for frame $i$.  
The final mSTIoU is obtained by averaging STIoU across all test clips.  
For ACC@0.5+n, a frame contributes a score of 1 if its IoU exceeds 0.5 (target-present), or if correctly classified as no-target. %; otherwise it contributes 0.

We compare VIRO against two types of baselines: (i) the fully supervised video REC model MDETR+BH trained on RefEgo, and (ii) off-the-shelf RefCOCOg-trained OFA~\citep{wang2022ofa} and MDETR~\citep{kamath2021mdetr} following the RefEgo setup.
As shown in Table~\ref{tab:refego}, VIRO demonstrates strong zero-shot capabilities.  
% As shown in Table~\ref{tab:refego}, VIRO demonstrates strong zero-shot capabilities.  
In the all-frames setting, including both target-present and no-target frames, VIRO attains competitive mSTIoU and surpasses the fully supervised method in ACC@0.5+n.  
Furthermore, it yields the highest performance among zero-shot methods on target-present frames, confirming its robust grounding ability without domain-specific training.

\subsection{Ablation Studies}\label{sec:exp:ablation}
In this section, we conduct ablation studies to analyze the impact of key hyperparameters on VIRO's performance and to validate our design choices. 
Unless stated otherwise, all experiments are conducted using VIRO with GroundingDINO-T on the RefCOCO testA split. Additional ablations, including analyses of the LLM backbones (llama3.1-Instruct~\cite{grattafiori2024llama}, GPT-4o/mini~\cite{hurst2024gpt}) and the early-exit mechanism, are provided in Appendix~\ref{appendix:ablations}. Furthermore, qualitative results and execution examples of our pipeline are detailed in Appendices~\ref{appendix:qualitative}~and~\ref{appendix:exp:VIRO_example}, respectively.

\vspace{-1em}
\paragraph{Verification Components.} 

Table~\ref{tab:ablation_verification} shows the results of a cumulative ablation study to analyze the contribution of each verification module of our VIRO pipeline. We start with a standard detector-only baseline and progressively add our proposed modules to measure the impact on both no-target robustness and standard REC accuracy.
Incorporating operators (without verification) boosts balanced accuracy to 56.8\%, suggesting that the compositional pipeline effectively grounds all query noun phrases.
Adding LV and UV further enhances no-target robustness, albeit with a slight TPR reduction reflecting a precision–recall trade-off.
% This improvement stems from the compositional pipeline's ability to ground all noun phrases in the query, enabling sophisticated reasoning.
% The addition of LV and UV yields a further incremental improvement in no-target robustness, with a small reduction in TPR that reflects a precision–recall trade-off.

\vspace{-1em}
\paragraph{OVD Detection Threshold.} The detection threshold of the Open-Vocabulary Detector (OVD) is a critical parameter that governs the trade-off between standard REC accuracy (TPR) and no-target robustness (TNR). As shown in Figure~\ref{fig:thres:detection}, a higher threshold improves TNR by filtering out spurious detections, but this simultaneously lowers recall, which in turn degrades TPR. We adopt a threshold of $0.2$ to favor high recall at the proposal stage while maintaining balanced overall performance.

\vspace{-1em}
\paragraph{CLIP Models.} 
We analyze the impact of the CLIP model backbone, used for both UV and the \texttt{PROPERTY} operator. Table~\ref{tab:ablation_clip} compares the performance of ViT-L/14 and ViT-H/14. Our final configuration uses ViT-H/14 for its optimal balance between accuracy and efficiency, while the lighter ViT-L/14 remains a computationally efficient alternative.

% --- Table 2: CLIP model ablation ---
\begin{table}[!htb]
  \centering
  \vspace{-0.3em}
  \caption{Ablation on the CLIP backbone, comparing the trade-off between TPR and execution-stage (Exec.) runtime.}
  \label{tab:ablation_clip}
  \vspace{-0.3em}
  \begin{adjustbox}{width=0.73 \linewidth}
    \begin{tabular}{lcc}
      \toprule
      {CLIP Model} & {TPR (Acc@0.5) $\uparrow$} & {Exec. $\downarrow$} \\
      \midrule
      ViT-H/14 & 71.9 & 0.71 \\ %1.39 \\
      ViT-L/14 & 68.8 & 0.55 \\ %1.79 \\
      \bottomrule
    \end{tabular}
  \end{adjustbox}
  \vspace{-1.0em}
\end{table}

\section{Discussion and Conclusion}\label{sec:conclusion}

In this work, we introduced VIRO, a verification-integrated neuro-symbolic pipeline for REC. By embedding lightweight verification mechanisms at each reasoning step, our framework explicitly handles no-target scenarios, supports interpretable early termination, and maintains state-of-the-art accuracy with efficient execution and favorable scalability.
Beyond REC, the modularity of our pipeline suggests promising extensions to interactive domains. In particular, the ability to parse natural language into verifiable symbolic programs opens the door to future applications where robots and humans engage in dialogue. In such scenarios, robots could not only interpret complex instructions but also execute them safely and transparently, ensuring that ambiguous commands are rejected before action. This direction underscores the broader potential of our approach as a foundation for trustworthy multimodal reasoning in embodied AI systems.

\vspace{-.8em}
\paragraph{Acknowledgements.}
This work was supported by Institute of Information \& Communications Technology Planning \& Evaluation (IITP) grants funded by the Korea government (MSIT) (RS-2019-II191906,Artificial Intelligence Graduate School Program (POSTECH); RS-2024-00457882, AI Research Hub Project; IITP-2026-RS-2024-00437866, Information Technology Research Center (ITRC)). It was also supported by the Korea Institute for Advancement of Technology (KIAT) grant funded by the Ministry of Trade, Industry and Energy (MOTIE) (RS-2025-00564342), and the Seoul R\&BD Program (SP240008) through the Seoul Business Agency (SBA) funded by the Seoul Metropolitan Government.

{
    \small
    \bibliographystyle{ieeenat_fullname}
    \bibliography{main}

@String(AAAI = {AAAI})

@inproceedings{han2024zero,
  title={Zero-shot referring expression comprehension via structural similarity between images and captions},
  author={Han, Zeyu and Zhu, Fangrui and Lao, Qianru and Jiang, Huaizu},
  booktitle={Proceedings of the IEEE/CVF Conference on Computer Vision and Pattern Recognition},
  pages={14364--14374},
  year={2024}
}

@article{chen2025flora,
  title={FLORA: Formal Language Model Enables Robust Training-free Zero-shot Object Referring Analysis},
  author={Chen, Zhe and Chen, Zijing},
  journal={arXiv preprint arXiv:2501.09887},
  year={2025}
}

@inproceedings{liu2024grounding,
  title={Grounding dino: Marrying dino with grounded pre-training for open-set object detection},
  author={Liu, Shilong and Zeng, Zhaoyang and Ren, Tianhe and Li, Feng and Zhang, Hao and Yang, Jie and Jiang, Qing and Li, Chunyuan and Yang, Jianwei and Su, Hang and others},
  booktitle={European Conference on Computer Vision},
  pages={38--55},
  year={2024},
  organization={Springer}
}

@article{he2023grec,
  title={GREC: Generalized referring expression comprehension},
  author={He, Shuting and Ding, Henghui and Liu, Chang and Jiang, Xudong},
  journal={arXiv preprint arXiv:2308.16182},
  year={2023}
}

@article{qiao2020referring,
  title={Referring expression comprehension: A survey of methods and datasets},
  author={Qiao, Yanyuan and Deng, Chaorui and Wu, Qi},
  journal={IEEE Transactions on Multimedia},
  volume={23},
  pages={4426--4440},
  year={2020},
  publisher={IEEE}
}

@inproceedings{wang2021structured,
  title={Structured scene memory for vision-language navigation},
  author={Wang, Hanqing and Wang, Wenguan and Liang, Wei and Xiong, Caiming and Shen, Jianbing},
  booktitle={Proceedings of the IEEE/CVF conference on Computer Vision and Pattern Recognition},
  pages={8455--8464},
  year={2021}
}

@inproceedings{zhang2024interactive,
  title={Interactive navigation in environments with traversable obstacles using large language and vision-language models},
  author={Zhang, Zhen and Lin, Anran and Wong, Chun Wai and Chu, Xiangyu and Dou, Qi and Au, KW Samuel},
  booktitle={2024 IEEE International Conference on Robotics and Automation},
  pages={7867--7873},
  year={2024},
  organization={IEEE}
}

@inproceedings{yu2018mattnet,
  title={Mattnet: Modular attention network for referring expression comprehension},
  author={Yu, Licheng and Lin, Zhe and Shen, Xiaohui and Yang, Jimei and Lu, Xin and Bansal, Mohit and Berg, Tamara L},
  booktitle={Proceedings of the IEEE Conference on Computer Vision and Pattern Recognition},
  pages={1307--1315},
  year={2018}
}

@inproceedings{kamath2021mdetr,
  title={Mdetr-modulated detection for end-to-end multi-modal understanding},
  author={Kamath, Aishwarya and Singh, Mannat and LeCun, Yann and Synnaeve, Gabriel and Misra, Ishan and Carion, Nicolas},
  booktitle={Proceedings of the IEEE/CVF International Conference on Computer Vision},
  pages={1780--1790},
  year={2021}
}

@inproceedings{yan2023universal,
  title={Universal instance perception as object discovery and retrieval},
  author={Yan, Bin and Jiang, Yi and Wu, Jiannan and Wang, Dong and Luo, Ping and Yuan, Zehuan and Lu, Huchuan},
  booktitle={Proceedings of the IEEE/CVF Conference on Computer Vision and Pattern Recognition},
  pages={15325--15336},
  year={2023}
}

@inproceedings{subramanian2022reclip,
  title={Reclip: A strong zero-shot baseline for referring expression comprehension},
  author={Subramanian, Sanjay and Merrill, William and Darrell, Trevor and Gardner, Matt and Singh, Sameer and Rohrbach, Anna},
  booktitle={Proceedings of the 60th Annual Meeting of the Association for Computational Linguistics (Volume 1: Long Papers)},
  pages={5198--5215},
  year={2022}
}

@inproceedings{shen2024groundvlp,
  title={Groundvlp: Harnessing zero-shot visual grounding from vision-language pre-training and open-vocabulary object detection},
  author={Shen, Haozhan and Zhao, Tiancheng and Zhu, Mingwei and Yin, Jianwei},
  booktitle={Proceedings of the AAAI Conference on Artificial Intelligence},
  volume={38},
  pages={4766--4775},
  year={2024}
}

@inproceedings{mao2016generation,
  title={Generation and comprehension of unambiguous object descriptions},
  author={Mao, Junhua and Huang, Jonathan and Toshev, Alexander and Camburu, Oana and Yuille, Alan L and Murphy, Kevin},
  booktitle={Proceedings of the IEEE Conference on Computer Vision and Pattern Recognition},
  pages={11--20},
  year={2016}
}

@inproceedings{yu2016modeling,
  title={Modeling context in referring expressions},
  author={Yu, Licheng and Poirson, Patrick and Yang, Shan and Berg, Alexander C and Berg, Tamara L},
  booktitle={European Conference on Computer Vision},
  pages={69--85},
  year={2016},
  organization={Springer}
}

@inproceedings{shridhar2022cliport,
  title={Cliport: What and where pathways for robotic manipulation},
  author={Shridhar, Mohit and Manuelli, Lucas and Fox, Dieter},
  booktitle={Conference on Robot Learning},
  pages={894--906},
  year={2022},
  organization={PMLR}
}

@article{jin2025referring,
  title={Referring Expression Comprehension in semi-structured human--robot interaction},
  author={Jin, Tianlei and Meng, Qiwei and Zhang, Gege and Huang, Qiulan and Guo, Fangtai and Kong, Shu and Song, Wei and Zhu, Jiakai and Gu, Jason},
  journal={Expert Systems with Applications},
  volume={275},
  pages={126965},
  year={2025},
  publisher={Elsevier}
}

@inproceedings{gupta2023visual,
  title={Visual programming: Compositional visual reasoning without training},
  author={Gupta, Tanmay and Kembhavi, Aniruddha},
  booktitle={Proceedings of the IEEE/CVF Conference on Computer Vision and Pattern Recognition},
  pages={14953--14962},
  year={2023}
}

@inproceedings{suris2023vipergpt,
  title={Vipergpt: Visual inference via python execution for reasoning},
  author={Sur{\'\i}s, D{\'\i}dac and Menon, Sachit and Vondrick, Carl},
  booktitle={Proceedings of the IEEE/CVF International Conference on Computer Vision},
  pages={11888--11898},
  year={2023}
}

@inproceedings{ke2024hydra,
  title={HYDRA: A Hyper Agent for Dynamic Compositional Visual Reasoning},
  author={Ke, Fucai and Cai, Zhixi and Jahangard, Simindokht and Wang, Weiqing and Haghighi, Pari Delir and Rezatofighi, Hamid},
  booktitle={European Conference on Computer Vision},
  pages={132--149},
  year={2024},
  organization={Springer}
}

@inproceedings{liu2023gres,
  title={Gres: Generalized referring expression segmentation},
  author={Liu, Chang and Ding, Henghui and Jiang, Xudong},
  booktitle={Proceedings of the IEEE/CVF Conference on Computer Vision and Pattern Recognition},
  pages={23592--23601},
  year={2023}
}

@inproceedings{xiao2024florence,
  title={Florence-2: Advancing a unified representation for a variety of vision tasks},
  author={Xiao, Bin and Wu, Haiping and Xu, Weijian and Dai, Xiyang and Hu, Houdong and Lu, Yumao and Zeng, Michael and Liu, Ce and Yuan, Lu},
  booktitle={Proceedings of the IEEE/CVF Conference on Computer Vision and Pattern Recognition},
  pages={4818--4829},
  year={2024}
}

@inproceedings{zeng2022multi,
  title={Multi-Grained Vision Language Pre-Training: Aligning Texts with Visual Concepts},
  author={Zeng, Yan and Zhang, Xinsong and Li, Hang},
  booktitle={International Conference on Machine Learning},
  pages={25994--26009},
  year={2022},
  organization={PMLR}
}

@inproceedings{cai2025naver,
  title     = {NAVER: A Neuro-Symbolic Compositional Automaton for Visual Grounding with Explicit Logic Reasoning},
  author    = {Cai, Zhixi and Ke, Fucai and Jahangard, Simindokht and Garcia de la Banda, Maria and Haffari, Reza and Stuckey, Peter J. and Rezatofighi, Hamid},
  booktitle = {Proceedings of the IEEE/CVF International Conference on Computer Vision},
  year      = {2025},
}

@inproceedings{li2022grounded,
  title={Grounded language-image pre-training},
  author={Li, Liunian Harold and Zhang, Pengchuan and Zhang, Haotian and Yang, Jianwei and Li, Chunyuan and Zhong, Yiwu and Wang, Lijuan and Yuan, Lu and Zhang, Lei and Hwang, Jenq-Neng and others},
  booktitle={Proceedings of the IEEE/CVF Conference on Computer Vision and Pattern Recognition},
  pages={10965--10975},
  year={2022}
}

@article{ren2016faster,
  title={Faster R-CNN: Towards real-time object detection with region proposal networks},
  author={Ren, Shaoqing and He, Kaiming and Girshick, Ross and Sun, Jian},
  journal={IEEE Transactions on Pattern Analysis and Machine Ontelligence},
  volume={39},
  number={6},
  pages={1137--1149},
  year={2016},
  publisher={IEEE}
}

@inproceedings{lin2014coco,
  title={Microsoft COCO: Common Objects in Context},
  author={Lin, Tsung-Yi and Maire, Michael and Belongie, Serge and Hays, James and Perona, Pietro and Ramanan, Deva and Doll{\'a}r, Piotr and Zitnick, C. Lawrence},
  booktitle={European Conference on Computer Vision},
  pages={740--755},
  year={2014},
  organization={Springer}
}

@article{qwen2.5,
      title={Qwen2 Technical Report}, 
      author={An Yang and others},
      journal={arXiv preprint arXiv:2407.10671},
      year={2024}
}

@inproceedings{yin2025unigoal,
  title={Unigoal: Towards universal zero-shot goal-oriented navigation},
  author={Yin, Hang and Xu, Xiuwei and Zhao, Linqing and Wang, Ziwei and Zhou, Jie and Lu, Jiwen},
  booktitle={Proceedings of the Computer Vision and Pattern Recognition Conference},
  pages={19057--19066},
  year={2025}
}

@inproceedings{yokoyama2024vlfm,
  title={Vlfm: Vision-language frontier maps for zero-shot semantic navigation},
  author={Yokoyama, Naoki and Ha, Sehoon and Batra, Dhruv and Wang, Jiuguang and Bucher, Bernadette},
  booktitle={2024 IEEE International Conference on Robotics and Automation (ICRA)},
  pages={42--48},
  year={2024},
  organization={IEEE}
}

@article{yang2024depth,
  title={Depth anything v2},
  author={Yang, Lihe and Kang, Bingyi and Huang, Zilong and Zhao, Zhen and Xu, Xiaogang and Feng, Jiashi and Zhao, Hengshuang},
  journal={Advances in Neural Information Processing Systems},
  volume={37},
  pages={21875--21911},
  year={2024}
}

@inproceedings{radford2021learning,
  title={Learning transferable visual models from natural language supervision},
  author={Radford, Alec and Kim, Jong Wook and Hallacy, Chris and Ramesh, Aditya and Goh, Gabriel and Agarwal, Sandhini and Sastry, Girish and Askell, Amanda and Mishkin, Pamela and Clark, Jack and others},
  booktitle={International Conference on Machine Learning},
  pages={8748--8763},
  year={2021},
  organization={PmLR}
}

@inproceedings{lee2024interactive,
  title={Interactive Text-to-Image Retrieval with Large Language Models: A Plug-and-Play Approach},
  author={Lee, Saehyung and Yu, Sangwon and Park, Junsung and Yi, Jihun and Yoon, Sungroh},
  booktitle={Proceedings of the 62nd Annual Meeting of the Association for Computational Linguistics (Volume 1: Long Papers)},
  pages={791--809},
  year={2024}
}

@inproceedings{deng2009imagenet,
  title={Imagenet: A large-scale hierarchical image database},
  author={Deng, Jia and Dong, Wei and Socher, Richard and Li, Li-Jia and Li, Kai and Fei-Fei, Li},
  booktitle={2009 IEEE conference on computer vision and pattern recognition},
  pages={248--255},
  year={2009},
  organization={Ieee}
}

@article{grattafiori2024llama,
  title={The llama 3 herd of models},
  author={Grattafiori, Aaron and Dubey, Abhimanyu and Jauhri, Abhinav and Pandey, Abhinav and Kadian, Abhishek and Al-Dahle, Ahmad and Letman, Aiesha and Mathur, Akhil and Schelten, Alan and Vaughan, Alex and others},
  journal={arXiv preprint arXiv:2407.21783},
  year={2024}
}

@article{chen2024far,
  title={How far are we to gpt-4v? closing the gap to commercial multimodal models with open-source suites},
  author={Chen, Zhe and Wang, Weiyun and Tian, Hao and Ye, Shenglong and Gao, Zhangwei and Cui, Erfei and Tong, Wenwen and Hu, Kongzhi and Luo, Jiapeng and Ma, Zheng and others},
  journal={Science China Information Sciences},
  volume={67},
  number={12},
  pages={220101},
  year={2024},
  publisher={Springer}
}

@inproceedings{akula2020words,
  title={Words aren’t enough, their order matters: On the robustness of grounding visual referring expressions},
  author={Akula, Arjun and Gella, Spandana and Al-Onaizan, Yaser and Zhu, Song-Chun and Reddy, Siva},
  booktitle={Proceedings of the 58th Annual Meeting of the Association for Computational Linguistics},
  pages={6555--6565},
  year={2020}
}

@inproceedings{kurita2023refego,
  title={Refego: Referring expression comprehension dataset from first-person perception of ego4d},
  author={Kurita, Shuhei and Katsura, Naoki and Onami, Eri},
  booktitle={Proceedings of the IEEE/CVF International Conference on Computer Vision},
  pages={15214--15224},
  year={2023}
}

@inproceedings{grauman2022ego4d,
  title={Ego4d: Around the world in 3,000 hours of egocentric video},
  author={Grauman, Kristen and Westbury, Andrew and Byrne, Eugene and Chavis, Zachary and Furnari, Antonino and Girdhar, Rohit and Hamburger, Jackson and Jiang, Hao and Liu, Miao and Liu, Xingyu and others},
  booktitle={Proceedings of the IEEE/CVF Conference on Computer Vision and Pattern Recognition},
  pages={18995--19012},
  year={2022}
}

@inproceedings{li2023evaluating,
  title={Evaluating object hallucination in large vision-language models},
  author={Li, Yifan and Du, Yifan and Zhou, Kun and Wang, Jinpeng and Zhao, Wayne Xin and Wen, Ji-Rong},
  booktitle={Proceedings of the 2023 Conference on Empirical Methods in Natural Language Processing},
  pages={292--305},
  year={2023}
}

@inproceedings{wang2022ofa,
  title={Ofa: Unifying architectures, tasks, and modalities through a simple sequence-to-sequence learning framework},
  author={Wang, Peng and Yang, An and Men, Rui and Lin, Junyang and Bai, Shuai and Li, Zhikang and Ma, Jianxin and Zhou, Chang and Zhou, Jingren and Yang, Hongxia},
  booktitle={International Conference on Machine Learning},
  pages={23318--23340},
  year={2022},
  organization={PMLR}
}

@inproceedings{li2023blip,
  title={Blip-2: Bootstrapping language-image pre-training with frozen image encoders and large language models},
  author={Li, Junnan and Li, Dongxu and Savarese, Silvio and Hoi, Steven},
  booktitle={International Conference on Machine Learning},
  pages={19730--19742},
  year={2023},
  organization={PMLR}
}

@inproceedings{ranftl2021vision,
  title={Vision transformers for dense prediction},
  author={Ranftl, Ren{\'e} and Bochkovskiy, Alexey and Koltun, Vladlen},
  booktitle={Proceedings of the IEEE/CVF International Conference on Computer Vision},
  pages={12179--12188},
  year={2021}
}

@inproceedings{kirillov2023segment,
  title={Segment anything},
  author={Kirillov, Alexander and Mintun, Eric and Ravi, Nikhila and Mao, Hanzi and Rolland, Chloe and Gustafson, Laura and Xiao, Tete and Whitehead, Spencer and Berg, Alexander C and Lo, Wan-Yen and others},
  booktitle={Proceedings of the IEEE/CVF International Conference on Computer Vision},
  pages={4015--4026},
  year={2023}
}

@article{bai2025qwen2,
  title={Qwen2.5-vl technical report},
  author={Bai, Shuai and Chen, Keqin and Liu, Xuejing and Wang, Jialin and Ge, Wenbin and Song, Sibo and Dang, Kai and Wang, Peng and Wang, Shijie and Tang, Jun and others},
  journal={arXiv preprint arXiv:2502.13923},
  year={2025}
}

@inproceedings{hudson2019gqa,
  title={Gqa: A new dataset for real-world visual reasoning and compositional question answering},
  author={Hudson, Drew A and Manning, Christopher D},
  booktitle={Proceedings of the IEEE/CVF Conference on Computer Vision and Pattern Recognition},
  pages={6700--6709},
  year={2019}
}

@article{hurst2024gpt,
  title={Gpt-4o system card},
  author={Hurst, Aaron and Lerer, Adam and Goucher, Adam P and Perelman, Adam and Ramesh, Aditya and Clark, Aidan and Ostrow, AJ and Welihinda, Akila and Hayes, Alan and Radford, Alec and others},
  journal={arXiv preprint arXiv:2410.21276},
  year={2024}
}
}

\appendix
\clearpage
\onecolumn 
\section{Appendix}

\renewcommand{\thefigure}{A\arabic{figure}}
\setcounter{figure}{0}
\renewcommand{\thetable}{A\arabic{table}}
\setcounter{table}{0}

We used an LLM as a general-purpose writing assistant for minor edits (clarity, grammar, or phrasing) and for generating alternative phrasings of paragraphs we had already drafted.
% The Use of Large Language Models (LLMs) \ck{check}
% The use of LLMs is allowed as a general-purpose assist tool. However, new this year, if LLMs played a significant role in research ideation and/or writing to the extent that they could be regarded as a contributor, then authors should describe the precise role of the LLM in a separate section on LLM usage. This section can appear in the appendix, and will not be considered as part of the page limit. Not disclosing significant LLM usage can lead to desk rejection of the paper.

\subsection{Details of VIRO}\label{appendix:operators}
\subsubsection{Input Arguments}
We provide the details of VIRO by extracting the program generator prompts. The following functions are available in our framework for reasoning:

\begin{itemize}
    \item \textbf{FIND(object\_name}=`object\_name'):
    Returns all objects matching the object name which are clearly detectable, excluding non-object entities (e.g., living room, field, wall).
    
    \item \textbf{LOCATE(object}=objects, \textbf{position}=`location'):
    Returns objects positioned at a specified absolute location, independent of other objects in the 2D space (e.g., `right', `at the bottom', `on left', `9 o clock', `outmost right', `top', `uppermost', `middle', `center').
    
    \item \textbf{ORDER(object}=objects, \textbf{criteria}=[`left'\textbar`right'\textbar`top'\textbar`bottom'], \textbf{rank}=number):
    Returns the object positioned at the specified \texttt{rank} when sorted only by the given criteria (`left', `right', `top', `bottom'), counting from the end.
    
    \item \textbf{ABSOLUTE\_DEPTH(object}=objects, \textbf{criteria}=[`front'\textbar`behind']):
    Returns objects from \texttt{objects} positioned absolutely closest (front) or farthest (behind) in the 3D space (depth information).
    
    \item \textbf{SIZE(object}=objects, \textbf{criteria}=[`big'\textbar`small']):
    Returns objects filtered by relative size only by the given criteria (`big', `small').
    
    \item \textbf{PROPERTY(object}=objects, \textbf{value}=`attribute'):
    Filters objects based on their intrinsic attributes (e.g., color and patterns: `red', `striped', clothing: `wearing a blue shirt', states or actions: `standing', `sitting', `turned on', `open').
    
    \item \textbf{FIND\_DIRECTION(object}=objects1, \textbf{reference\_object}=objects2, \textbf{direction}=[`left'\textbar`right'\textbar`top'\textbar`bottom']):
    Returns objects from \texttt{objects1} positioned next to objects in \texttt{objects2} only by the given criteria (`left', `right', `top', `bottom').
    
    \item \textbf{FIND\_NEAR(object}=objects1, \textbf{reference\_object}=objects2):
    Returns objects from \texttt{objects1} that are spatially close to any object in \texttt{objects2}.
    
    \item \textbf{FIND\_INSIDE(object}=objects1, \textbf{reference\_object}=objects2):
    Returns objects from \texttt{objects1} that are strictly inside the reference object \texttt{objects2}.
    
    \item \textbf{RELATIVE\_DEPTH(object}=objects1, \textbf{reference\_object}=objects2, \textbf{criteria}=[`front'\textbar`behind']):
    Returns objects from \texttt{objects1} positioned in depth relative to objects in \texttt{objects2} only by the given criteria (`front', `behind').
    
    \item \textbf{RESULT(object}=answer\_object):
    Pre-processes the final selected object to the final answer form.
\end{itemize}

\subsubsection{Verification Module}
\noindent\textbf{Attribute Operator.} 
We use the \texttt{PROPERTY} operator to filter visual attributes using the CLIP and GroundingDINO-T model. CLIP has varying thresholds depending on the given image, which makes filtering based solely on similarity scores challenging. To improve the filtering process, we first apply a softmax transformation to the CLIP similarity scores of the candidate regions from the OVD. We then integrate GroundingDINO-T scores into the filtering process, combining them with the softmax CLIP scores via a weighted sum. Finally, we set an adaptive threshold based on the number of candidates from \texttt{FIND} operator.

\smallskip
\noindent\textbf{Relative Spatial Operators.}
Relative spatial operators take at least two arguments: \texttt{object} and \texttt{reference\_object}. 
The \texttt{reference\_object} serves as the spatial standard, and the \texttt{object} denotes the entity whose position is specified relative to this reference.
As discussed in Section~\ref{sec:method:vro}, we use logical verification in \texttt{RELATIVE\_DEPTH}, similar to \texttt{FIND\_DIRECTION}, but with the relative depth values `front' or `behind'. Additionally, in \texttt{FIND\_INSIDE}, if there is no intersection area between the \texttt{object} and \texttt{reference\_object}, the proposal is rejected.

\subsubsection{Adaptive Threshold}\label{appendix:threshold}

To compute the filtering score $S(l|I_j)$ in equation~\eqref{eq:score}, for each candidate region $I_j$ with label $l$, we leverage a pre-trained VLM trained with contrastive loss, such as CLIP. CLIP is designed to align visual and textual representations in a shared embedding space, enabling effective discrimination between semantically relevant (positive) and irrelevant (negative) pairs.
However, one of the main challenges when using CLIP for verification is its inherent bias toward labels that appear frequently in its training data. Labels like `person' or `car' typically receive higher confidence scores compared to more specialized or rare objects, making a fixed threshold inappropriate for fair evaluation across different object categories.
To address this label-specific bias, we implement an adaptive thresholding mechanism that calibrates the decision boundary for each target label individually. We utilize ImageNet as an auxiliary calibration dataset, computing verification scores for a representative sample of images containing various object categories. 

We collect 5 images per class in ImageNet (a total of 5,000 images) and use GroundingDINO to crop out only the relevant class objects (denoted as $D_A)$. This process helps minimize bias from background elements, where many images contain a person even when the target class is not `person'.
For each target label $l$, we analyze the distribution of verification scores $S(l|I)$ across this preprocessed dataset to determine CLIP's typical confidence range for that specific category.

The calibration process employs a top-$k$ selection strategy to determine the appropriate threshold $\delta_l$
Specifically, we rank all verification scores computed on the auxiliary dataset for target label $l$ where $d\in D_A$, and the score is defined as:
\begin{align}
    S(l\,|\,d_i)=\frac{1}{K}\sum_{k=1}^{K}
    \frac{\exp(\operatorname{sim}(d_i,l)/\tau)}
    {\exp(\operatorname{sim}(d_i,l)/\tau)+\exp(\operatorname{sim}(d_i,c_k)/\tau)} \;.
\end{align}
The set of verification scores for the entire dataset is represented as:
\begin{align}
    \mathcal{S}(l|D_A)=\{S(l|d_1), S(l|d_2), \dots, S(l|d_n)\} \;,
\end{align}
where $n$ is the number of auxiliary dataset.
The threshold $\delta_l$ selects the top-$k$\% highest scoring samples, where $k$ is a hyperparameter (typically set as 10).
This strategy effectively captures the performance level that CLIP consistently achieves for high-confidence predictions of the target category, providing a data-driven threshold that reflects CLIP's inherent capability for that specific label.

\begin{figure}
    \centering
    \includegraphics[width=0.32\linewidth]{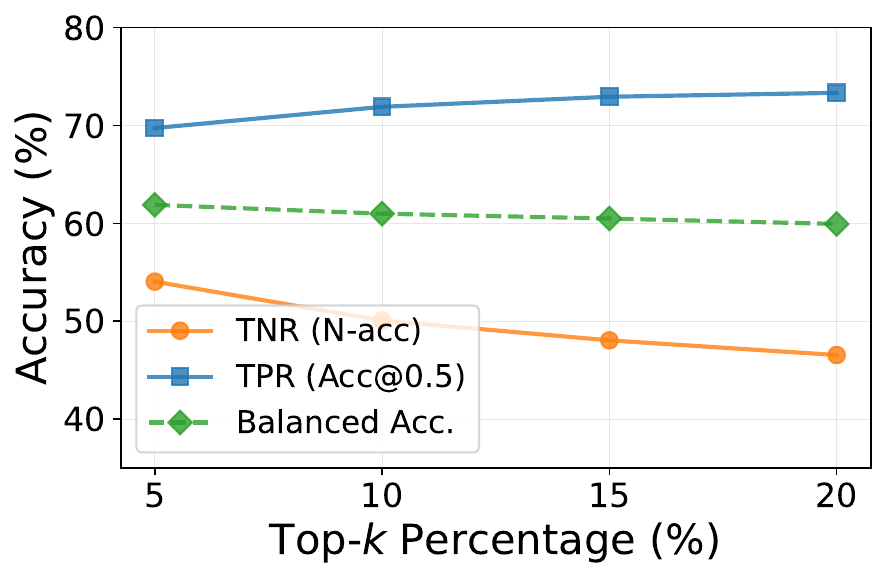}
    \caption{Analysis of $k$ adaptive threshold which illustrates the trade-off between TPR and TNR.}
    \vspace{-1em}
    \label{fig:k_ablation}
\end{figure}

We also investigate the sensitivity of our pipeline to $k$. As shown in Figure~\ref{fig:k_ablation}, VIRO’s performance remains highly robust to changes in the Top-$k$ percentage. Although a slight increase in TPR is observed with more candidates (as the ground-truth box is more likely to be included), the overall Balanced Accuracy stays stable across different values of $k$.

\subsubsection{Justification for Neuro-symbolic Verification}
VIRO addresses two fundamental limitations in existing approaches: \textit{unreliable verification} and \textit{cascading errors}.
First, free-form Python approaches rely on LLM-generated code for verification (\textit{e.g.}, \texttt{if} blocks), which is not explicitly enforced and thus prone to being incorrectly generated or omitted. In contrast, VIRO restricts program generation to a pre-defined set of verifiable primitives.
Furthermore, methods like NAVER perform verification only 
after all reasoning steps, making them prone to 
cascading errors: inevitable false positives from open-vocabulary 
detectors propagate through intermediate reasoning 
steps and corrupt final predictions.
In contrast, VIRO’s operator-level verification ensures high-precision grounding by filtering out incorrect candidates.

\subsection{Program Validator}\label{appendix:validator}

The Program Validator serves as a safeguard against errors in LLM-generated programs. Its primary purpose is to detect syntactic, structural, and logical mistakes before program execution, and to provide structured feedback tailored to the specific type of error. Since LLM outputs are not guaranteed to be flawless, the validator detects invalid programs at pre-execution time and provides feedback that guides the LLM to regenerate a correct program.

A valid program must follow a simple and strict format: each line is written as 
\texttt{VAR = OP(ARG=..., ...)} and executed sequentially in order, while the program always terminates with a line that must take the form 
\texttt{FINAL\_RESULT = RESULT(object=VAR)}. Given this constrained design, the validator enforces the following properties:
\begin{itemize}
    \item \textbf{Syntax enforcement:} all lines except the last must take the form \texttt{VAR = OP(ARG=..., ...)}, and the last line must take the form \texttt{FINAL\_RESULT = RESULT(object=VAR)}.
    \item \textbf{Variable tracking:} at line $t$, only variables defined in lines $1,\dots,t-1$ may be referenced, and redefinition of an existing variable is disallowed to prevent overwriting outputs from earlier steps.
    \item \textbf{Argument typing:} arguments are checked for valid type (variable, string, or number). Certain arguments have additional constraints, e.g., \texttt{rank} must be a positive integer and \texttt{criteria} must be chosen from a predefined set (e.g., \texttt{left}, \texttt{right}, \texttt{top}, \texttt{bottom}).
    \item \textbf{Operator constraints:} each operator must belong to the predefined functions (e.g., \texttt{FIND}, \texttt{PROPERTY}), and each function must include all of its required arguments.
    \item \textbf{Output format:} the program must end with a single line in the form \texttt{FINAL\_RESULT = RESULT(object=VAR)}, and no other \texttt{RESULT} may appear earlier in the program.
\end{itemize}

Through these checks, the Program Validator acts as a reliable feedback mechanism, reducing execution errors caused by incorrect code generated by the LLM.
Both our method and ViperGPT~\cite{suris2023vipergpt} use programs generated by the LLM prior to execution, but the scope of validation differs fundamentally.
ViperGPT only ensures that the generated code is syntactically valid Python code, which allows issues such as undefined variables, incorrect argument types, or missing required arguments to pass through unnoticed until runtime.
In contrast, our validator applies strict structural and semantic checks before execution, ensuring correct format, valid variables, and a proper output statement.
These checks reduce execution errors and provide targeted feedback for correction.

% \subsection{Validator Percent}\label{appendix:validator}
\subsection{Experimental Setup}\label{appendix:exp}
% : $0.2$ for GroundingDINO-T and $0.5$ for GLIP-L. 

\subsubsection{Datasets}\label{appendix:exp:datasets}
\noindent\textbf{RefCOCO/+/g.} RefCOCO~\citep{yu2016modeling} primarily targets location-based expressions, 
while RefCOCO+~\citep{yu2016modeling} focuses on attribute-based descriptions by prohibiting the use of absolute location words. 
RefCOCOg~\citep{mao2016generation}, in contrast, contains longer and more complex expressions, often combining both spatial relations and attributes. 
For RefCOCO and RefCOCO+, results are reported separately on two test splits: TestA, where the referent is a person, and TestB, where the referent is a non-person object, highlighting object-centric grounding. For RefCOCOg, results are reported on the standard single test split.
All of these benchmarks are built upon the MSCOCO~\citep{lin2014coco} image dataset.

\smallskip
\noindent\textbf{gRefCOCO no-target split.} 
% \paragraph{gRefCOCO.}
gRefCOCO~\citep{he2023grec, liu2023gres} is a referring expression dataset that explicitly includes no-target queries, allowing evaluation of systems that must either localize the referred region or abstain when no valid target exists. 
To avoid trivial negatives, no-target expressions are constrained to be contextually related to the image, and annotators may reuse deceptive expressions from the same split when needed. 

\smallskip
\noindent\textbf{RefAdv.} RefAdv~\citep{akula2020words} is an out-of-distribution (OOD) adversarial test set derived from Ref-Hard, the structure-dependent subset of RefCOCOg, designed to evaluate whether referring expression models truly leverage syntactic structure rather than rely on lexical cues. It reveals models’ lack of syntactic understanding and limited generalization, even for those that perform well on the original RefCOCOg benchmark.

% Ref-Hard is the structure-dependent subset of RefCOCOg, constructed by selecting examples that become difficult for humans when the word order is shuffled. For each such case, the authors chose as the distractor (new target) the object that humans most frequently confused when the word order was altered. Crowd-workers then rewrote the original expressions—keeping most content words unchanged but modifying syntax and word order—so that the new expressions referred to the distractor. Each rewritten item was included only if at least two out of three annotators could correctly localize the new target (IoU > 0.5). RefAdv thus serves to reveal models’ lack of syntactic understanding and limited generalization, even for those performing well on the original RefCOCOg benchmark.}

\smallskip
\noindent\textbf{RefEgo.}
RefEgo~\citep{kurita2023refego} is a video-based referring expression comprehension (REC) dataset built upon first-person videos from the Ego4D~\citep{grauman2022ego4d}, designed to evaluate models’ ability to localize objects described in natural language under realistic conditions. The dataset contains clips, which are selected to feature rapid egocentric camera motion and multiple similar objects, and some include frames where the target object is absent, requiring models to detect no-target situations. Unlike prior REC datasets built from web-curated images, RefEgo focuses on testing language–vision grounding and generalization in real-world environments.

\subsubsection{Evaluation Metrics}
We assess both \textit{no-target robustness} and standard REC accuracy, which jointly require addressing localization and classification. 
Following a binary classification view, we define outcomes based on the confusion matrix:
\textbf{True Positive (TP)}: a target is present and the model correctly localizes it (Intersection-over-Union (IoU) $>$ 0.5);  
\textbf{True Negative (TN)}: target is absent and the model correctly predicts its absence;  
\textbf{False Positive (FP)}: target is absent but the model incorrectly outputs a bounding box; and  
\textbf{False Negative (FN)}: a target is present but the model either predicts `no-target' or localizes it incorrectly (IoU $<$ 0.5).

\subsubsection{Baseline Details}\label{appendix:exp:baseline}
\noindent\textbf{Fully supervised REC} baselines are trained with full annotations on RefCOCO/+/g. 
GREC~\citep{he2023grec} further incorporates the gRefCOCO no-target dataset to address target-absent cases. 
By contrast, Qwen2.5-VL-72B-AWQ~\cite{bai2025qwen2} is evaluated without fine-tuning on gRefCOCO. 
To enable no-target prediction, we append a negative instruction to its prompt, \textit{e.g.}, ``if there is no object, return []'', thereby leveraging its inherent reasoning capability. 
The full prompt for Qwen2.5-VL is as follows: Detect \{description\} in the image. Return a JSON list of bounding boxes as [x1, y1, x2, y2] in pixel coordinates relative to the input image. If there is no object, return [].

\noindent\textbf{Proposal-based REC} first parses the referring expression to isolate key linguistic components before matching them against pre-generated object proposals. \textbf{ReCLIP}~\citep{subramanian2022reclip} employs a syntactic parser to extract noun chunks, while \textbf{GroundVLP}~\citep{shen2024groundvlp} utilizes a traditional NLP toolbox to identify the main object. More recent approaches leverage the advanced capabilities of Large Language Models (LLMs) for this task; \textbf{SS-CLIP/SS-FLAVA}~\citep{han2024zero}, two variants of the method in \citep{han2024zero}, use an LLM to parse the main object from the query. After this initial parsing stage, each method employs its unique mechanism to map the extracted components to the most relevant regions in given detected proposals from MAttNet~\citep{yu2018mattnet}. 
These are architecturally constrained to a rich pool of candidate regions by Faster RCNN, making a direct comparison on target-absent task unfair.
% For fair comparison, we isolate the linguistic parser from each baseline and feed its output—the identified main object—directly into GroundingDINO.

\smallskip
\noindent\textbf{Compositional reasoning REC} parses complex queries into explicit programs. By generating and then executing these programs, they transparently handle multi-step compositional logic to derive the final result. We compare our approach with state-of-the-art methods in this domain, including \textbf{ViperGPT} \citep{suris2023vipergpt}, \textbf{HYDRA} \citep{ke2024hydra}, and \textbf{NAVER} \citep{cai2025naver}, to benchmark its compositional reasoning capabilities.

%Additional hyperparameters and program generation prompts are provided in Appendix~\ref{appendix:exp:implementation}.

\subsection{Standard REC Accuracy Under Forced Prediction}\label{appendix:forced_choice}

Our framework is evaluated under a target-present assumption, wherein each query is guaranteed to have a corresponding target in the image. In this setting, our approach achieves high accuracy on standard benchmarks, as summarized in Table~\ref{tab:forced_choice}.

\begin{table*}[htb!]
\centering
\caption{Comparison of REC methods on standard benchmarks in terms of TPR (Acc@0.5), under a forced prediction setting.}
\begin{adjustbox}{width=0.68\linewidth}
\begin{tabular}{lcccccccc}
\toprule
& \multicolumn{3}{c}{RefCOCO} & 
\multicolumn{3}{c}{RefCOCO+} & \multicolumn{2}{c}{RefCOCOg} \\
\cmidrule(lr){2-4} \cmidrule(lr){5-7} \cmidrule(lr){8-9}
Method & Val & TestA & TestB & Val & TestA & TestB & Val & Test \\
\midrule
\multicolumn{5}{l}{\textbf{\textit{Fully Supervised REC}}} \\
Qwen2.5-VL-72B~\citep{bai2025qwen2} & 92.7 & 94.6 & 89.7 & 88.9 & 92.2 & 83.7 & 89.9 & 90.3 \\
GREC-MDETR-R101~\citep{kamath2021mdetr} & 86.8 & 89.6 & 81.4 & 79.5 &	84.1 & 70.6 & 81.6 & 80.9 \\
GREC-UNINEXT-R50~\citep{yan2023universal} & 89.7 & 91.5 & 86.9 & 79.8 & 85.2 & 72.8 & 84.0 & 84.3 \\
\midrule
\multicolumn{5}{l}{\textbf{\textit{Proposal-based REC}}} \\
ReCLIP~\citep{subramanian2022reclip} & 45.8 & 47.0 & 45.2 & 45.3 & 48.5 & 42.7 & 57.0 & 56.2 \\
SS-CLIP~\citep{han2024zero} & 60.6 & 66.5 & 54.9 & 55.5 & 62.6 & 45.7 & 59.9 & 59.9 \\
SS-FLAVA~\citep{han2024zero} & 52.5 & 52.7 & 52.9 & 50.8 & 53.4 & 47.6 & 61.3 & 60.9 \\
GroundVLP~\citep{shen2024groundvlp} & 52.6 & 61.3 & 43.5 & 56.4 & 64.8 & 47.4 & 64.3 & 63.5 \\

\midrule
\multicolumn{5}{l}{\textit{\textbf{Detector-based REC}}} \\
GLIP-L~\citep{li2022grounded} & 47.5 &	52.6 &	41.8 & 44.1 & 48.6 & 39.8 & 51.9 & 52.6 \\
GroundingDINO-T~\citep{liu2024grounding} & 50.4 & 57.2 & 43.2 & 51.4 & 57.6 & 45.8 & 60.4 & 59.5 \\ 

\midrule
\multicolumn{5}{l}
{\textit{\textbf{Compositional Reasoning REC with GLIP}}} \\
ViperGPT~\citep{suris2023vipergpt} & 66.9 & 72.0 & 59.9 & 59.6 & 65.7 & 63.0 & 69.3 & 69.6 \\
HYDRA~\citep{ke2024hydra} & 68.0 & 73.1 & 62.5 & 55.8 & 60.6 & 50.6 & 67.2 & 67.6 \\
NAVER~\citep{cai2025naver} & 69.6 & 73.4 & 64.4 & 59.0 & 62.7  & 56.4 & 70.7 & 70.0 \\
\cellcolor[HTML]{EFEFEF}\textbf{VIRO (Ours)} &
\cellcolor[HTML]{EFEFEF}71.4 &
\cellcolor[HTML]{EFEFEF}75.7 & 
\cellcolor[HTML]{EFEFEF}63.8 & 
\cellcolor[HTML]{EFEFEF}59.3 & 
\cellcolor[HTML]{EFEFEF}66.2 & 
\cellcolor[HTML]{EFEFEF}51.3 & 
\cellcolor[HTML]{EFEFEF}70.6 & 
\cellcolor[HTML]{EFEFEF}71.5 \\

\midrule
\multicolumn{5}{l}
{\textit{\textbf{Compositional Reasoning REC with GroundingDINO}}} \\
ViperGPT~\citep{suris2023vipergpt} & 62.2 & 66.7 & 54.6 & 55.4 & 61.7 & 50.4	& 66.0 & 65.7 \\
HYDRA~\citep{ke2024hydra} & 62.0 & 62.8 & 62.4 & 55.0 & 58.4 & 51.6 & 66.4 & 67.1  \\
NAVER~\citep{cai2025naver} & 61.1 & 64.2 & 58.2 & 56.4 & 60.1 & 51.8 & 68.4 & 68.4 \\
GDINO-FLORA$^{\ddagger}$~\citep{chen2025flora} & 73.7 & 78.5 & 67.8 & 63.2 & 71.6 & 53.5 & 72.5 & 72.1 \\
\cellcolor[HTML]{EFEFEF}\textbf{VIRO (Ours)} &
\cellcolor[HTML]{EFEFEF}71.4 &
\cellcolor[HTML]{EFEFEF}75.0 &
\cellcolor[HTML]{EFEFEF}64.4 &
\cellcolor[HTML]{EFEFEF}59.5 &
\cellcolor[HTML]{EFEFEF}65.8 &
\cellcolor[HTML]{EFEFEF}50.1 &
\cellcolor[HTML]{EFEFEF}69.6 &
\cellcolor[HTML]{EFEFEF}70.3 \\
\bottomrule
\multicolumn{9}{l}{\footnotesize $^{\ddagger}$ Official code has not been released as of September 25, 2025.} \\
\end{tabular}
\end{adjustbox}
\label{tab:forced_choice}
\end{table*}

\subsection{Handling No-Target Cases and Errors in Baselines}\label{appendix:explanation_notarget}

This section defines how we count no-target predictions and attribute errors for the baselines evaluated in our study.
For \textbf{detector-based baselines},
if the detector returns no bounding boxes for the query, we label the prediction as no-target (correct). 
For \textbf{compositional-reasoning REC} baselines, prior implementations do not explicitly consider no-target cases; we therefore specify our handling of no-target outcomes and errors. Specifically, if the framework’s final output is the empty set, we count the prediction as \emph{no-target (correct)}. Further method-specific details are provided below.

\smallskip
\noindent\textbf{ViperGPT~\citep{suris2023vipergpt}.}
In our evaluation of ViperGPT, an instance is classified as a no-target if the final code execution resulted in an empty list. Errors are identified when the \texttt{exec(compile(code, 'Codex', 'exec'), globals())} call returned None, which typically signifies a failure during code execution. ViperGPT’s instructions enforce a fallback that retrieves the entire image when an object is not detected (e.g., when \texttt{len(object)==0}). This design choice creates ambiguity in identifying no-targets case. While likely intended to prevent downstream Python errors, this execution-safety behavior makes it difficult to distinguish a true `no object found' scenario. We therefore adopt a conservative rule: treat only an empty final result as no-target, and treat None as error.

\smallskip
\noindent\textbf{HYDRA~\citep{ke2024hydra}.}
For HYDRA, we adapt a strict criterion for no-target cases: we label no-target (correct) only when the number of detected main objects or relation-forming objects is exactly zero (\textit{i.e.}, \texttt{len(...)==0}). 
Conversely, any case with \texttt{len(...)$\ge$ 1} is classified as an error. 
%An output of len(...) >= 1 introduces ambiguity. 
While it could potentially represent a no-target scenario (\textit{e.g.}, the model detected objects but failed to establish the queried relationship), it is not possible to definitively confirm this. Therefore, to maintain a conservative count of true no-target instances, we opt to classify these ambiguous cases as errors.

% The rationale for this decision is twofold:
% An output of len(...) == 0 provides a high-confidence confirmation that the model identified no-targets.

\smallskip
\noindent\textbf{NAVER~\citep{cai2025naver}.}
The NAVER pipeline operates through four distinct stages, each equipped with a self-correction mechanism: Perception, Logic Generation, Logic Reasoning, and Logic Answering. If the Perception stage fails to produce region proposals, it retries with a lower object detection threshold. In the Logic Generation stage, a failure prompts the LLM to retry the generation process. If the Logic Reasoning stage encounters a code error or yields no-target candidates, the pipeline reverts to the Logic Generation stage to create a new logical program. Finally, in the Answering stage, a Multimodal Large Language Model (MLLM)—specifically InternVL2-8B~\citep{chen2024far}—is prompted with the question, ``Does the object meet the query?" If the MLLM answers `no,' this stage is also retried.

After a maximum of five retry attempts across the pipeline, a query is classified as a no-target under one of three conditions:
(i) perception failure (no detection), (ii) logic reasoning fails to establish the required relations (e.g., \texttt{contains}, \texttt{inside}, \texttt{is}), and (iii) the MLLM consistently answers `no,' in answering state, indicating that no identified object satisfies the query.

Based on our implementation of the official code (at the time of our experiments), we observe that the vast majority of processing errors occur during the Logic Reasoning stage, specifically during the translation to or execution of the Probabilistic Problog program. Further analysis of NAVER's no-target cases is provided in Appendix~\ref{appendix:naver}.

\subsection{Detailed Experimental Results}\label{appendix:all_results}

We provide full benchmark results and baseline details across all evaluation splits for completeness.
% For completeness, this section presents the full performance and explanation of baselines across all evaluation splits of the benchmark datasets.

\subsubsection{Baseline Implementation for No-Target Cases}\label{appendix:no-target}

We evaluate our model's ability to handle negative cases where the target is not present in the image. The results on the gRefCOCO no-target split are shown in Table~\ref{tab:gref}.

\begin{table*}[htb!]
\centering
\caption{Quantitative evaluation of no-target robustness on the gRefCOCO no-target split, reported as TNR (N-acc) (\%).} %(Acc@0.5) (\%).}
\begin{adjustbox}{width=.49\linewidth}
\begin{tabular}{lccc}
\toprule
Method & Val & TestA & TestB \\
\midrule
\multicolumn{3}{l}{\textbf{\textit{Fully Supervised REC}}} \\
GREC-MDETR-R101~\citep{kamath2021mdetr} & 36.3 & 34.5 & 31.0 \\
GREC-UNINEXT-R50~\citep{yan2023universal} & 50.6 & 49.3 & 48.2 \\
\midrule
\multicolumn{3}{l}{\textbf{\textit{Detector-based REC}}} \\
GLIP-L~\citep{li2022grounded} & 14.8 & 21.7 & 18.2 \\
GroundingDINO-T~\citep{liu2024grounding} &  2.0 & 22.8 & 16.0 \\ 
\midrule
\multicolumn{3}{l}{\textbf{\textit{Compositional Reasoning REC with GLIP}}} \\
ViperGPT~\citep{suris2023vipergpt} & 0.3 & 0.1 & 0.1 \\
HYDRA~\citep{ke2024hydra} & 7.2 & 6.3 & 6.1 \\
NAVER~\citep{cai2025naver} & 14.9 & 13.9 & 13.0 \\
\cellcolor[HTML]{EFEFEF}\textbf{VIRO (Ours)} &
\cellcolor[HTML]{EFEFEF}56.7 & \cellcolor[HTML]{EFEFEF}50.1 & \cellcolor[HTML]{EFEFEF}53.2 
\\
\midrule
\multicolumn{3}{l}{\textbf{\textit{Compositional Reasoning REC with GroundingDINO}}} \\
ViperGPT~\citep{suris2023vipergpt} & 
0.3 & 0.2 &	0.1 \\
HYDRA~\citep{ke2024hydra} & 8.6 & 7.5 & 7.0 \\
NAVER~\citep{cai2025naver} & 3.0 & 3.4 & 1.8 \\
\cellcolor[HTML]{EFEFEF}\textbf{VIRO (Ours)} &
\cellcolor[HTML]{EFEFEF}56.5 & \cellcolor[HTML]{EFEFEF}50.2 &	\cellcolor[HTML]{EFEFEF}52.9 \\
\bottomrule
\end{tabular}
\end{adjustbox}
\label{tab:gref}
\end{table*}

\subsubsection{Baseline Implementation for Standard REC Benchmarks}\label{appendix:standard}

The detailed results for the RefCOCO, RefCOCO+, and RefCOCOg datasets are provided in Table~\ref{tab:refcoco}, Table~\ref{tab:refcoco+}, and Table~\ref{tab:refcocog}, respectively.

\begin{table*}[htb!]
\centering
\caption{Quantitative results on the RefCOCO dataset. Accuracy (\%) is evaluated under two conditions: including all predictions (Inc.$\uparrow$) and excluding model failure cases (Exc.$\uparrow$). FR denotes the overall Failure Rate (\%).}
\begin{adjustbox}{width=0.75\linewidth}
\begin{tabular}{lccccccccc}
\toprule
& \multicolumn{3}{c}{Val} & \multicolumn{3}{c}{TestA} & \multicolumn{3}{c}{TestB} \\
\cmidrule(lr){2-4}\cmidrule(lr){5-7}\cmidrule(lr){8-10}
Method & FR $\downarrow$ & Exc.$\uparrow$ & Inc.$\uparrow$ & FR $\downarrow$ & Exc.$\uparrow$ & Inc.$\uparrow$ & FR $\downarrow$ & Exc.$\uparrow$ & Inc.$\uparrow$ \\
\midrule
\multicolumn{5}{l}{\textbf{\textit{Detector-based REC}}} \\
GLIP-L~\citep{li2022grounded} & 0.00 & 47.5 & 47.5 & 0.00 & 52.6 & 52.6 & 0.00  & 41.8 & 41.8 \\
GroundingDINO-T~\citep{liu2024grounding} & 0.00 & 50.4 & 50.4 & 0.00 & 57.2 & 57.2 & 0.00 & 43.2 & 43.2 \\ 
% GLIP-L
\midrule
\multicolumn{5}{l}{\textbf{\textit{Compositional Reasoning REC with GLIP}}} \\
ViperGPT~\citep{suris2023vipergpt} & 4.12 & 66.9 & 64.1 & 3.32 & 72.0 & 69.6 & 5.16 & 59.9 & 56.8 \\
HYDRA~\citep{ke2024hydra} & 19.60 & 68.0 & 54.7 & 17.27 & 73.1 & 60.5 & 24.14 & 62.5 & 47.4  \\
NAVER~\citep{cai2025naver} & 17.97 & 69.6 & 57.1 & 14.85 & 73.4 & 62.5 & 20.71 & 64.4 & 51.1  \\
\cellcolor[HTML]{EFEFEF}\textbf{VIRO (Ours)} &
\cellcolor[HTML]{EFEFEF}0.09 & \cellcolor[HTML]{EFEFEF}68.1 & \cellcolor[HTML]{EFEFEF}68.1 & \cellcolor[HTML]{EFEFEF}0.07 & \cellcolor[HTML]{EFEFEF}72.8 & \cellcolor[HTML]{EFEFEF}72.8 & \cellcolor[HTML]{EFEFEF}0.14 & \cellcolor[HTML]{EFEFEF}60.6 & \cellcolor[HTML]{EFEFEF}60.5 \\
\midrule
\multicolumn{5}{l}{\textbf{\textit{Compositional Reasoning REC with GroundingDINO}}} \\
ViperGPT~\citep{suris2023vipergpt} & 3.90 & 62.2 & 59.8 & 3.45 & 66.7 & 64.4 & 4.71 & 54.6 & 52.0 \\
HYDRA~\citep{ke2024hydra} & 26.13 & 62.0 & 45.8 & 28.46 & 62.8 & 44.9 & 34.6 & 62.4 & 40.8 \\
NAVER~\citep{cai2025naver} & 8.77 & 61.1 & 55.7 & 6.01 & 64.2 & 60.3 & 10.70 & 58.2 & 52.0 \\
\cellcolor[HTML]{EFEFEF}\textbf{VIRO (Ours)} &
\cellcolor[HTML]{EFEFEF}0.09 & \cellcolor[HTML]{EFEFEF}68.2 & \cellcolor[HTML]{EFEFEF}68.1 & \cellcolor[HTML]{EFEFEF}0.07 & \cellcolor[HTML]{EFEFEF}71.9 & \cellcolor[HTML]{EFEFEF}71.9 & \cellcolor[HTML]{EFEFEF}0.14 & \cellcolor[HTML]{EFEFEF}60.8 & \cellcolor[HTML]{EFEFEF}60.8 \\
\bottomrule
\end{tabular}
\end{adjustbox}
\label{tab:refcoco}
\end{table*}

\begin{table*}[htb!]
\centering
\caption{Quantitative results on the RefCOCO+ dataset. Accuracy (\%) is evaluated under two conditions: including all predictions (Inc.$\uparrow$) and excluding model failure cases (Exc.$\uparrow$). FR denotes the overall Failure Rate (\%).}
\begin{adjustbox}{width=0.75\linewidth}
\begin{tabular}{lccccccccc}
\toprule
& \multicolumn{3}{c}{Val} & \multicolumn{3}{c}{TestA} & \multicolumn{3}{c}{TestB} \\
\cmidrule(lr){2-4}\cmidrule(lr){5-7}\cmidrule(lr){8-10}
Method & FR $\downarrow$ & Exc.$\uparrow$ & Inc.$\uparrow$ & FR $\downarrow$ & Exc.$\uparrow$ & Inc.$\uparrow$ & FR $\downarrow$ & Exc.$\uparrow$ & Inc.$\uparrow$ \\
\midrule
\multicolumn{5}{l}{\textbf{\textit{Detector-based REC}}} \\
GLIP-L~\citep{li2022grounded} & 0.00 & 44.1 & 44.1 & 0.00 & 48.6 & 48.6 & 0.00 & 39.8 & 39.8 \\
GroundingDINO-T~\citep{liu2024grounding} & 0.00 & 51.4 & 51.4 & 0.00 & 57.6 & 57.6 & 0.00 & 45.8 & 45.8 \\ 
\midrule
\multicolumn{5}{l}{\textbf{\textit{Compositional Reasoning REC with GLIP}}} \\
ViperGPT~\citep{suris2023vipergpt} & 6.97 & 59.6 & 55.4 & 3.28 & 65.7 & 63.6 & 6.73 & 53.0 & 49.4 \\
HYDRA~\citep{ke2024hydra} & 24.40 & 55.8 & 42.2 & 17.06 & 60.6 & 50.3 & 30.91 & 50.6 & 35.0 \\
NAVER~\citep{cai2025naver} & 26.44 & 59.0 & 43.4 & 14.67 & 62.7 & 53.5 & 33.59 & 56.4 & 37.5 \\
\cellcolor[HTML]{EFEFEF}\textbf{VIRO (Ours)} &
\cellcolor[HTML]{EFEFEF}0.08 & \cellcolor[HTML]{EFEFEF}56.1 & \cellcolor[HTML]{EFEFEF}56.0 &
\cellcolor[HTML]{EFEFEF}0.07 & \cellcolor[HTML]{EFEFEF}63.7 & \cellcolor[HTML]{EFEFEF}63.7 &
\cellcolor[HTML]{EFEFEF}0.12 & \cellcolor[HTML]{EFEFEF}47.0 & \cellcolor[HTML]{EFEFEF}47.0 \\
\midrule
\multicolumn{5}{l}{\textbf{\textit{Compositional Reasoning REC with GroundingDINO}}} \\
ViperGPT~\citep{suris2023vipergpt} & 6.71 & 55.4 & 51.7 & 6.83 & 61.7 & 57.5 & 6.46 & 50.4 & 47.1 \\
HYDRA~\citep{ke2024hydra} & 33.16 & 55.0 & 36.8 & 35.96 & 58.4 & 37.4 & 29.72 & 51.6 & 36.3 \\
NAVER~\citep{cai2025naver} & 14.50 & 56.4 & 48.2 & 7.39 & 60.1 & 55.7 & 20.82 & 51.8 & 41.0 \\
\cellcolor[HTML]{EFEFEF}\textbf{VIRO (Ours)} &
\cellcolor[HTML]{EFEFEF}0.08 & \cellcolor[HTML]{EFEFEF}56.6 & \cellcolor[HTML]{EFEFEF}56.5 &
\cellcolor[HTML]{EFEFEF}0.00 & \cellcolor[HTML]{EFEFEF}63.3 & \cellcolor[HTML]{EFEFEF}63.3 &
\cellcolor[HTML]{EFEFEF}0.12 & \cellcolor[HTML]{EFEFEF}46.2 & \cellcolor[HTML]{EFEFEF}46.2 \\
\bottomrule
\end{tabular}
\end{adjustbox}
\label{tab:refcoco+}
\end{table*}

\begin{table*}[htb!]
\centering
\caption{Quantitative results on the RefCOCOg dataset. Accuracy (\%) is evaluated under two conditions: including all predictions (Inc.$\uparrow$) and excluding model failure cases (Exc.$\uparrow$). FR denotes the overall Failure Rate (\%).}
\begin{adjustbox}{width=0.57\linewidth}
\begin{tabular}{lcccccc}
\toprule
& \multicolumn{3}{c}{Val} & \multicolumn{3}{c}{Test}  \\
\cmidrule(lr){2-4}\cmidrule(lr){5-7}
Method & FR $\downarrow$ & Exc.$\uparrow$ & Inc.$\uparrow$ & FR $\downarrow$ & Exc.$\uparrow$ & Inc.$\uparrow$ \\
\midrule
\multicolumn{5}{l}{\textbf{\textit{Detector-based REC}}} \\
GLIP-L~\citep{li2022grounded} & 0.00 & 51.9 & 51.9 & 0.00 & 52.6 & 52.6 \\
GroundingDINO-T~\citep{liu2024grounding} & 0.00 & 60.4 & 60.4 & 0.00 & 59.5 & 59.5 \\ 
\midrule
\multicolumn{5}{l}{\textbf{\textit{Compositional Reasoning REC with GLIP}}} \\
ViperGPT~\citep{suris2023vipergpt} & 7.40 & 69.3 & 64.2 & 6.31 & 69.6 & 65.2 \\
HYDRA~\citep{ke2024hydra} & 23.98 & 67.2 & 51.1 & 21.44 & 67.6 & 53.1 \\
NAVER~\citep{cai2025naver} & 28.44 & 70.7 & 50.6 & 25.90 & 70.0 & 51.9 \\
\cellcolor[HTML]{EFEFEF}\textbf{VIRO (Ours)} &
\cellcolor[HTML]{EFEFEF}0.28 & \cellcolor[HTML]{EFEFEF}66.2 & \cellcolor[HTML]{EFEFEF}66.0 &
\cellcolor[HTML]{EFEFEF}0.38 & \cellcolor[HTML]{EFEFEF}67.0 & \cellcolor[HTML]{EFEFEF}66.8
\\
\midrule
\multicolumn{5}{l}{\textbf{\textit{Compositional Reasoning REC with GroundingDINO}}} \\
ViperGPT~\citep{suris2023vipergpt} & 6.55 & 66.0 & 61.7 & 6.03 & 65.7 & 61.7 \\
HYDRA~\citep{ke2024hydra} & 34.07 & 66.4 & 43.8 & 32.37 & 67.1 & 45.4 \\
NAVER~\citep{cai2025naver} & 38.50 & 68.4 & 42.1 & 9.74 & 68.4 & 55.8 \\
\cellcolor[HTML]{EFEFEF}\textbf{VIRO (Ours)} &
\cellcolor[HTML]{EFEFEF}0.24 & \cellcolor[HTML]{EFEFEF}66.2 & \cellcolor[HTML]{EFEFEF}66.0 &
\cellcolor[HTML]{EFEFEF}0.38 & \cellcolor[HTML]{EFEFEF}66.6 & \cellcolor[HTML]{EFEFEF}66.3 \\
\bottomrule
\end{tabular}
\end{adjustbox}
\label{tab:refcocog}
\end{table*}

\subsubsection{Analysis of NAVER}\label{appendix:naver} 

The Perception, Logic Reasoning, and Logic Answering modules in NAVER invoke self-correction when no valid target is found, effectively forcing an object prediction. To evaluate performance on the no-target split, we disable this behavior in all three modules and adopt an \emph{early-exit} policy that stops the pipeline when ``no-target'' is detected. We report TNR on the gRefCOCO testA no-target set, and TPR on the RefCOCO testA set. As shown in Table~\ref{tab:naver_ablation}, disabling self-correction improves TNR but at the cost of TPR. 
The module-wise distribution of early exits---indicating which component identifies the target's absence)---is summarized in Table~\ref{tab:naver_frequency}. 
This analysis is intended to isolate no-target behavior and does not reflect the default official setting. %Notice that \emph{Logic Answering} employs InternVL2~\citep{chen2024far}, which is strong but incurs substantial execution-time overhead.

\begin{table*}[htb!]
\centering
\caption{Impact of the self-correction on the NAVER framework. Performance is evaluated on gRefCOCO no-target TestA (TNR) and RefCOCO TestA (TPR) datasets.}
\begin{adjustbox}{width=0.6\linewidth}
\begin{tabular}{lccc}
\toprule
 % & & No Target Robustness & Standard REC \\
\cmidrule(lr){3-3} \cmidrule(lr){4-4}
Method & Balanced Acc $\uparrow$ & TNR {\scriptsize (gRef)} $\uparrow$ & TPR {\scriptsize (Ref)} $\uparrow$ \\
\midrule
NAVER w/ self-correction   & 33.8 & 3.4 & 64.2 \\
NAVER w/o self-correction  & 63.2 & 71.6 & 54.8 \\
% VIRO (Ours)                & 61.1 & 50.2 & 49.8 & 71.9 \\
\bottomrule
\end{tabular}
\end{adjustbox}
\label{tab:naver_ablation}
\end{table*}

\begin{table*}[htb!]
\centering
\caption{Frequency (\%) of early exits per NAVER module contributing to TNR on the gRefCOCO no-target TestA split.}
\begin{adjustbox}{width=0.3\linewidth}
\begin{tabular}{lc}
\toprule
\textbf{Module} & \textbf{Frequency (\%)} \\
\midrule
Perception & 67.8 \\
Logic Reasoning & 11.7 \\
Logic Answering & 20.5 \\
\bottomrule
\end{tabular}
\end{adjustbox}
\label{tab:naver_frequency}
\end{table*}

\subsubsection{Results on RefAdv dataset}\label{appendix:refadv}

To further evaluate the robustness of VIRO to linguistic perturbations, we analyze performance on the RefAdv benchmark, which introduces adversarially reordered referring expressions designed to test whether models rely on true syntactic understanding rather than superficial lexical cues. As shown in Table~\ref{tab:refadv}, VIRO achieves 64.2\% Exc. and 63.8\% Inc. accuracy on RefAdv, substantially outperforming prior compositional baselines, whose accuracy drops sharply due to frequent program-generation or execution failures.

\begin{table*}[htb!]
\centering
\caption{Quantitative results with GroundingDINO on the RefAdv dataset. Accuracy (\%) is evaluated under two conditions: including all dataset (Inc.$\uparrow$) and excluding model failure cases (Exc.$\uparrow$). FR denotes the overall Failure Rate (\%).}
\begin{adjustbox}{width=0.4\linewidth}
\begin{tabular}{lccc}
\toprule
& \multicolumn{3}{c}{Test}  \\
\cmidrule(lr){2-4}
Method & FR $\downarrow$ & Exc. $\uparrow$ & Inc. $\uparrow$ \\
\midrule
\multicolumn{4}{l}{\textbf{\textit{Detector-based REC}}} \\
GLIP-L~\citep{li2022grounded} & 0.00 & 55.7 & 55.7 \\
GroundingDINO-T~\citep{liu2024grounding} & 0.00 & 60.5 & 60.5 \\ 
\midrule
\multicolumn{4}{l}{\textbf{\textit{Compositional Reasoning REC}}} \\
ViperGPT~\citep{suris2023vipergpt} & 11.47 & 64.8 & 57.3 \\
HYDRA~\citep{ke2024hydra} & 36.61 & 65.8 & 41.7 \\
NAVER~\citep{cai2025naver} & 13.23 & 64.0  & 55.5 \\
\cellcolor[HTML]{EFEFEF}\textbf{VIRO (Ours)} &
\cellcolor[HTML]{EFEFEF}0.59 & \cellcolor[HTML]{EFEFEF}64.2 & \cellcolor[HTML]{EFEFEF}63.8 \\
\bottomrule
\end{tabular}
\end{adjustbox}
\label{tab:refadv}
\end{table*}

% \subsection{Heuristic}\label{appendix:heuristic_notarget}

% \subsubsection{Analysis of No-Target Decisions of Compositional Reasoning REC Approaches}

% \subsection{Parser of Implicit Decomposition Approaches}\label{appendix:parser} > 시간 되면 정리

% \subsection{More details of Semantic Clarification}\label{appendix:sc}
% \begin{itemize}
%     \item prompt of LLM
%     \item prompt of VLM
%     \item the specific (technical details) of fits for RefCOCO dataset (spcific for test A)
% \end{itemize}

\subsubsection{Results on RefEgo validation set}\label{appendix:refego_val}

In this section, we provide a detailed performance analysis on the RefEgo validation set in Table~\ref{tab:refego_val}, extending the test set results presented in the main paper.
While VIRO is not explicitly trained on egocentric video data, it significantly outperforms other off-the-shelf REC models such as OFA and MDETR.
Notably, VIRO achieves 49.6\% in Acc@0.5+n, demonstrating its strong zero-shot transferability in complex grounding scenarios without any task-specific fine-tuning.

\begin{table}[htb!]
\centering
\caption{Comparison of video-based REC performance on the RefEgo validation set.}%Quantitative results on the RefEgo dataset. Accuracy (\%) is evaluated under two conditions: including all predictions (Inc.$\uparrow$) and excluding model failure cases (Exc.$\uparrow$). FR denotes the overall Failure Rate (\%).}
\vspace{-.3em}
\begin{adjustbox}{width=.5\linewidth}
\begin{tabular}{lccc}
\toprule
& \multicolumn{2}{c}{All Frames} & Target-present \\%multicolumn{2}{c}{target-present}  \\
\cmidrule(lr){2-3}\cmidrule(lr){4-4}
Method & mSTIoU & Acc@0.5+n & TPR (Acc@0.5) \\
\midrule
\multicolumn{4}{l}{\textbf{\textit{Fully Supervised with RefEgo}}} \\
MDETR+BH~\cite{kurita2023refego} & 37.9 & 51.9 & 52.9 \\
\midrule
\multicolumn{4}{l}{\textbf{\textit{Off-the-shelf RefCOCOg model}}} \\
OFA~\cite{wang2022ofa} & 16.9 & 30.8 & 29.6 \\
MDETR~\cite{kamath2021mdetr} & 17.4 & 28.3 & 25.2 \\
\midrule
\multicolumn{4}{l}{\textbf{\textit{Compositional Reasoning REC with GroundingDINO}}} \\
ViperGPT~\cite{suris2023vipergpt} & 9.5 & 16.4 & 14.3 \\
\rowcolor[HTML]{EFEFEF}\textbf{VIRO (Ours)} & 23.0 & 49.6 & 33.9 \\
\bottomrule
\end{tabular}
\end{adjustbox}
\label{tab:refego_val}
\end{table}

\subsection{Runtime Analysis with LLM backbones}\label{appendix:runtime}

% We report end-to-end runtime per query decomposed into two stages: \emph{pre-execution} and \emph{execution}. 
% \textbf{Pre-execution} covers all processing prior to the first image-dependent computation, while \textbf{execution} comprises all subsequent image-involving runtime.
% Table~\ref{tab:ablation_pre_exe} reports the per-query totals on the RefCOCO testA split evaluated on an NVIDIA RTX A6000 GPU.

We analyze the end-to-end latency of VIRO by decomposing the per-query runtime into two distinct stages: \emph{pre-execution} and \emph{execution}.
As shown in Table~\ref{tab:ablation_pre_exe}, 
API-based models significantly reduce pre-execution overhead while maintaining competitive TPR.
While a large-scale local LLM (Qwen2.5-72B-Instruct-AWQ) requires 12.21s for pre-execution, API-based acceleration via GPT-4o reduces this latency to 1.06s---a nearly 11.5× speedup—while even achieving higher TPR (72.3\%). Consequently, the total runtime becomes dominated solely by the execution stage (fixed at 0.71s), rather than the reasoning overhead. This efficiency is a direct consequence of VIRO's decoupled design, which enables flexible backbone selection to optimize latency-accuracy trade-offs.

\begin{table*}[h!]
\centering
\centering
\caption{
    Pre-execution (Pre-exec.) runtime and accuracy comparison across different LLM backbones on RefCOCO testA. 
    \textbf{The execution stage maintains a constant latency of 0.71s across all models}, reflecting VIRO's decoupled architecture. 
    % API-based models significantly reduce pre-execution overhead while maintaining competitive TPR.
}
\begin{adjustbox}{width=0.52\columnwidth}
\begin{tabular}{lcc}
\toprule
\textbf{LLM Backbone} & \textbf{Pre-exec. (s)} & \textbf{TPR (\%)} \\
\midrule
Llama3.1-8B-Instruct (Local)     & 2.07  & 66.3 \\
Qwen2.5-72B-Instruct-AWQ (Local) & 12.21 & 71.9 \\
GPT-4o mini (API)                & 1.37  & 69.1 \\
GPT-4o (API)                     & 1.06  & 72.3 \\
\bottomrule
\end{tabular}
\end{adjustbox}
\label{tab:ablation_pre_exe}
\end{table*}

\subsection{Models Used in Compositional Baselines and VRAM Usage}\label{appendix:models}

Table~\ref{tab:vlm_comparison} details the pre-trained vision-language components employed by each compositional baseline and their corresponding peak VRAM consumption, measured on an NVIDIA RTX A6000 (48GB). To ensure a focused comparison on the vision-language backbone efficiency, the reported memory excludes the VRAM occupied by the LLM itself.

Existing compositional baselines typically rely on heavy Multimodal LLMs (MLLMs), which incur significant memory and computational overhead. To avoid such overhead, VIRO is designed to operate without a MLLM, instead utilizing a more lightweight verification process. Furthermore, unlike NAVER, which requires an additional SegmentAnything~\citep{kirillov2023segment} model, VIRO maintains a minimal set of components, requiring only 9.7GB of VRAM.
Note that VRAM usage among baselines may vary depending on specific model-loading strategies, such as full pre-loading versus modular execution.

\begin{table}[h!]
\centering
\caption{Comparison of visual-language components and peak VRAM usage. Memory is measured during inference, excluding the LLM backbone.}
\small
\renewcommand{\arraystretch}{1.1}
\begin{tabular}{
    l
    >{\centering\arraybackslash}p{3.0cm}
    >{\centering\arraybackslash}p{3.0cm}
    >{\centering\arraybackslash}p{3.0cm}
    >{\centering\arraybackslash}p{3.0cm}
}
\toprule
 & \textbf{ViperGPT}~\citep{suris2023vipergpt} & \textbf{HYDRA}~\citep{ke2024hydra} & \textbf{NAVER}~\citep{cai2025naver} & \textbf{VIRO (Ours)} \\
\midrule
\textbf{Multimodal Encoder} & X-VLM~\citep{zeng2022multi} & X-VLM~\citep{zeng2022multi} & X-VLM~\citep{zeng2022multi} & CLIP~\citep{radford2021learning} \\
\textbf{Open-set Detector} & GDINO-T~\citep{liu2024grounding} & GDINO-T~\citep{liu2024grounding} & GDINO-T~\citep{liu2024grounding} & GDINO-T~\citep{liu2024grounding} \\
\textbf{Depth Estimator} & DPT-Large~\citep{ranftl2021vision} & DPT-Large~\citep{ranftl2021vision} & DepthAnythingV2~\citep{yang2024depth} & DepthAnythingV2~\citep{yang2024depth} \\
\textbf{Multimodal LLM} & BLIP-2~\citep{li2023blip} & BLIP-2~\citep{li2023blip} & InternVL2~\citep{chen2024far} & N/A \\
\textbf{Mask Generator} & N/A & N/A & SegmentAnything~\citep{kirillov2023segment} & N/A \\
\midrule
\textbf{VLM Memory Load} & 35.0 GB  & 21.8GB & 29.9 GB & 9.7 GB \\
\bottomrule
\end{tabular}
\label{tab:vlm_comparison}
\end{table}

\subsection{Further Ablation Analysis}\label{appendix:ablations}
\subsubsection{LLM Ablation Studies}\label{appendix:llm}

To demonstrate that VIRO's performance stems from its architectural design rather than a specific LLM, we evaluate its robustness and scalability across various backbones, ranging from the Llama-3.1-8B~\cite{grattafiori2024llama} to the GPT-4o/mini~\cite{hurst2024gpt}.

First, we conduct an ablation study in which we replace the primary LLM backbone used in our main experiments with Llama-3.1-8B-Instruct~\cite{grattafiori2024llama}, while keeping all other components and hyperparameters unchanged. The results, presented in Table~\ref{tab:llama3.1-8B}, show that VIRO achieves a lower program-generation failure rate and maintains strong performance with Llama-3.1-8B. In contrast, ViperGPT and NAVER exhibit very high failure rates, which leads to substantially reduced Inc. TPR. We omit the results for HYDRA with Llama-3.1-8B-Instruct, as the model failed on nearly all queries; specifically, it struggled to adhere to HYDRA’s iterative generation instructions, where even a minor formatting error caused the entire execution pipeline to collapse.

\begin{table*}[htb!]
\centering
\caption{Performance comparison on standard REC benchmarks using Llama3.1-8B-Instruct~\cite{grattafiori2024llama} as an alternative LLM backbone. We report results on the testA splits of RefCOCO/+ and the test split of RefCOCOg. Accuracy is measured both including (Inc.$\uparrow$) and excluding (Exc.$\uparrow$) failure cases. The failure rate (\%) is reported for RefCOCOg.}
\vspace{-.5em}
\begin{adjustbox}{width=.7\linewidth}
\begin{tabular}{lccccccc}
\toprule
 & & \multicolumn{2}{c}{RefCOCO} & \multicolumn{2}{c}{RefCOCO+} & \multicolumn{2}{c}{RefCOCOg}  \\
 \cmidrule(lr){3-4} \cmidrule(lr){5-6} \cmidrule(lr){7-8}
Method & Failure Rate $\downarrow$ & Exc. $\uparrow$ & Inc. $\uparrow$ & Exc. $\uparrow$ & Inc. $\uparrow$ & Exc. $\uparrow$ & Inc. $\uparrow$ \\ % & FPS $\uparrow$ \\
\midrule
\multicolumn{5}{l}{\textbf{\textit{Detector-based REC}}} \\
GLIP-L~\citep{li2022grounded} & 0.00 & 52.6 & 52.6 & 48.6 & 48.6 & 52.6 & 52.6 \\
GroundingDINO-T~\citep{liu2024grounding} & 0.00 & 57.2 & 57.2 & 57.6 & 57.6 & 59.5 & 59.5 \\
\midrule
\multicolumn{5}{l}{\textbf{\textit{Compositional Reasoning REC with GLIP}}} \\
ViperGPT~\citep{suris2023vipergpt} & 40.09 & 65.1 & 48.4 & 53.2 & 41.3 & 57.6 & 35.1 \\
% HYDRA~\citep{ke2024hydra} & - & - & - & - & - & - & - \\
NAVER~\citep{cai2025naver} & 70.15 & 72.3 & 32.4 & 65.1 & 28.6 & 69.6 & 20.8 \\
\cellcolor[HTML]{EFEFEF}\textbf{VIRO (Ours)} & \cellcolor[HTML]{EFEFEF}0.34 & \cellcolor[HTML]{EFEFEF}68.2 & \cellcolor[HTML]{EFEFEF}68.2 & \cellcolor[HTML]{EFEFEF}56.6 & \cellcolor[HTML]{EFEFEF}56.6 & \cellcolor[HTML]{EFEFEF}61.4 & \cellcolor[HTML]{EFEFEF}61.4 \\
\midrule
\multicolumn{5}{l}{\textbf{\textit{Compositional Reasoning REC with GroundingDINO}}} \\
ViperGPT~\citep{suris2023vipergpt} & 41.82 & 59.1 & 43.7 & 48.0 & 37.5 & 52.1 & 31.7 \\
% HYDRA~\citep{ke2024hydra} & - & - & - & - & - & - & - \\
NAVER~\citep{cai2025naver} & 54.84 & 61.3 & 43.1 & 59.7 & 40.9 & 64.7 & 29.2 \\
\cellcolor[HTML]{EFEFEF}\textbf{VIRO (Ours)} & \cellcolor[HTML]{EFEFEF}0.34 & \cellcolor[HTML]{EFEFEF}66.3 & \cellcolor[HTML]{EFEFEF}66.3 & \cellcolor[HTML]{EFEFEF}55.3 & \cellcolor[HTML]{EFEFEF}55.3 & \cellcolor[HTML]{EFEFEF}60.9 & \cellcolor[HTML]{EFEFEF}60.9 \\
\bottomrule
\end{tabular}
\end{adjustbox}
\label{tab:llama3.1-8B}
\end{table*}

We further explore the scalability of our framework using GPT-4o and GPT-4o mini, while simultaneously clarifying the performance gap observed with the official NAVER results. As shown in Table~\ref{tab:gpt4o-mini}, the high performance of NAVER (91.7\%) stems from their use of a GroundingDINO-B detector fine-tuned on RefCOCO. To ensure a strictly zero-shot evaluation, we instead utilize GroundingDINO-T, which has not been exposed to the RefCOCO training set.
Notably, when evaluated using the same zero-shot detector, VIRO consistently outperforms NAVER, confirming that our framework provides stronger zero-shot performance.

\begin{table}[htb!]
\centering
\caption{Comparison of performance on the RefCOCO testA split using GPT-4o and GPT-4o mini. For a fair comparison, all methods are evaluated in a zero-shot setting except for the official NAVER entry using a fine-tuned detector.}
\label{tab:gpt4o-mini}
\begin{adjustbox}{width=0.54\linewidth}
\begin{tabular}{lccc}
\toprule
Method & LLM & Detector & Acc@0.5 \\
\midrule
NAVER (Official) & GPT-4o-mini & GroundingDINO-B$^*$ & 91.7 \\
\midrule
\multicolumn{4}{l}{\textbf{\textit{Fair Comparison: Zero-shot Setting}}} \\
NAVER (Official) & GPT-4o-mini & GroundingDINO-T & 64.9 \\
\rowcolor[HTML]{EFEFEF}\textbf{VIRO (Ours)} & GPT-4o-mini & GroundingDINO-T & 69.1 \\
\rowcolor[HTML]{EFEFEF}\textbf{VIRO (Ours)} & GPT-4o & GroundingDINO-T & \textbf{72.3} \\
% \rowcolor[HTML]{EFEFEF}& GPT-4o-mini & GroundingDINO-T & 69.1 \\
% \rowcolor[HTML]{EFEFEF}\multirow{-2}{*}{\textbf{VIRO (Ours)}} & GPT-4o & GroundingDINO-T & \textbf{72.3} \\
\bottomrule
\multicolumn{4}{l}{\footnotesize $^*$Fine-tuned on the RefCOCO training set.} \\
\end{tabular}
\end{adjustbox}
\end{table}

\subsubsection{Early-exit Mechanism.} % Our pipeline's early-exit mechanism enhances computational efficiency by halting execution as soon as any verification step fails. 
As shown in Table~\ref{tab:ablation_ee}, incorporating the early-exit reduces the average latency focusing on the program execution phase to 0.52 seconds per query on the gRefCOCO no-target testA split.
Since operators run sequentially, unmet intermediate conditions trigger immediate termination of the execution pipeline. For example, in the query "an elephant to the left of a man," the program exits as soon as the elephant is not found, thereby pruning redundant downstream visual operations. This mechanism effectively boosts throughput in scenarios where no-target cases are frequent.

% --- Table 1: Early-exit ablation ---
\begin{table}[htb!]
  \centering
  \caption{Ablation study on the early-exit mechanism. Latency and FPS are measured solely for the program execution stage, evaluated on the gRefCOCO no-target testA split.}
  \label{tab:ablation_ee}
  \vspace{-0.3em}
  \begin{adjustbox}{width=0.26\linewidth}
    \begin{tabular}{l S[table-format=1.2] S[table-format=1.2]}
      \toprule
      {Early-exit} & {Latency $\downarrow$} & {FPS $\uparrow$} \\
      \midrule
      Enabled  & 0.52 & 1.92 \\
      Disabled & 0.58 & 1.79 \\
      \bottomrule
    \end{tabular}
  \end{adjustbox}
\end{table}

% % --- Table 2: CLIP model ablation ---
% \begin{table}[t]
%   \centering
%   \caption{Ablation study on the CLIP model.}
%   \label{tab:ablation_clip}
%   \vspace{-0.3em}
%   \begin{adjustbox}{max width=\linewidth}
%     \begin{tabular}{l S[table-format=2.1] S[table-format=1.2]}
%       \toprule
%       {CLIP Model} & {TPR (Acc@0.5) $\uparrow$} & {FPS $\uparrow$} \\
%       \midrule
%       ViT-H/14 & 71.9 & 1.39 \\
%       ViT-L/14 & 68.8 & 1.79 \\
%       \bottomrule
%     \end{tabular}
%   \end{adjustbox}
%   \vspace{-0.5em}
% \end{table}

\subsection{Extension to Visual Question Answering via GQA}\label{appendix:gqa}
% To demonstrate VIRO's generalization beyond REC, we extend our framework to the GQA dataset~\cite{hudson2019gqa} (compositional VQA) in Table~\ref{tab:gqa}. 
As a preliminary extension beyond REC, we adapt VIRO to the GQA dataset~\cite{hudson2019gqa}, a compositional VQA benchmark, as shown in Table~\ref{tab:gqa}. For a fair comparison, we utilize the same backbone models (BLIP-2, GLIP, and GPT-3.5-turbo) following the setup in HYDRA.
For this task, we introduce two new operators: \texttt{ASK} for QA generation and \texttt{CROP\_DIRECTION} for spatial focusing.
This adaptation is implemented modularly by (i) defining the corresponding Python logic and (ii) registering the operators in the in-context prompt, allowing VIRO to support this task without retraining.
% Crucially, this extension is achieved via a modular process: (i) defining the Python logic, and (ii) registering it to the in-context prompt, which allows VIRO to adapt to new tasks without retraining.

\begin{table}[htb!]
\centering
\caption{Comparison of accuracy (\%) on the GQA dataset.}
\label{tab:gqa}
\begin{adjustbox}{width=0.22\linewidth}
\begin{tabular}{lc}
\toprule
Method & Acc \\
\midrule
\multicolumn{2}{l}{\textbf{\textit{Trained RL Controller}}} \\
HYDRA (Official) & 47.9 \\
\midrule
\multicolumn{2}{l}{\textbf{\textit{Zero-shot Setting}}} \\
ViperGPT (Official) & 37.9 \\
\rowcolor[HTML]{EFEFEF} \textbf{VIRO (Ours)} & \textbf{45.4} \\
\bottomrule
\end{tabular}
\end{adjustbox}
\end{table}

\subsection{Qualitative Analysis}\label{appendix:qualitative}
\begin{figure}[htb!]
    \centering
    \includegraphics[width=0.85\linewidth]{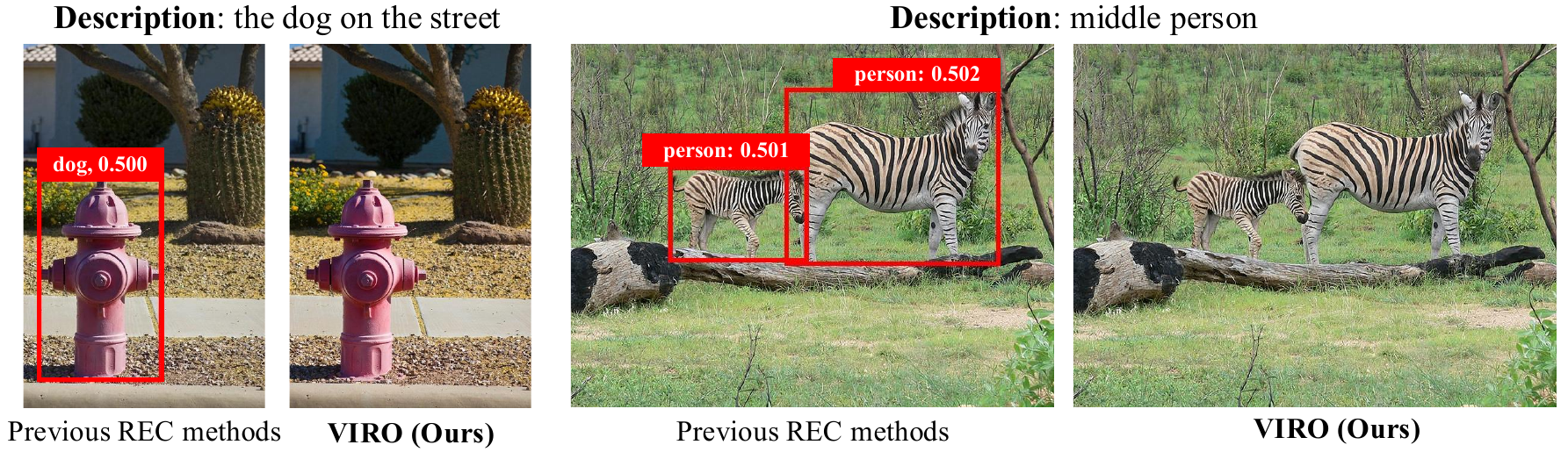}
    \caption{RefCOCO validation examples for false-positive suppression. Prior REC methods (left) yield spurious detections (red), whereas VIRO (right) rejects them via CLIP-based verification.}
    \label{fig:qual:FP}
  \label{fig:ovd+qual_row}
\end{figure}

We present qualitative examples from RefCOCO validation set. 
Figure~\ref{fig:qual:FP} demonstrates VIRO’s ability to suppress false positives (FPs) from open-vocabulary detectors via CLIP-based verification, compared against previous REC methods.

\subsection{VIRO Execution Examples}\label{appendix:exp:VIRO_example}
\begin{figure*}[htb!]
    \centering
    \includegraphics[width=.75\textwidth]{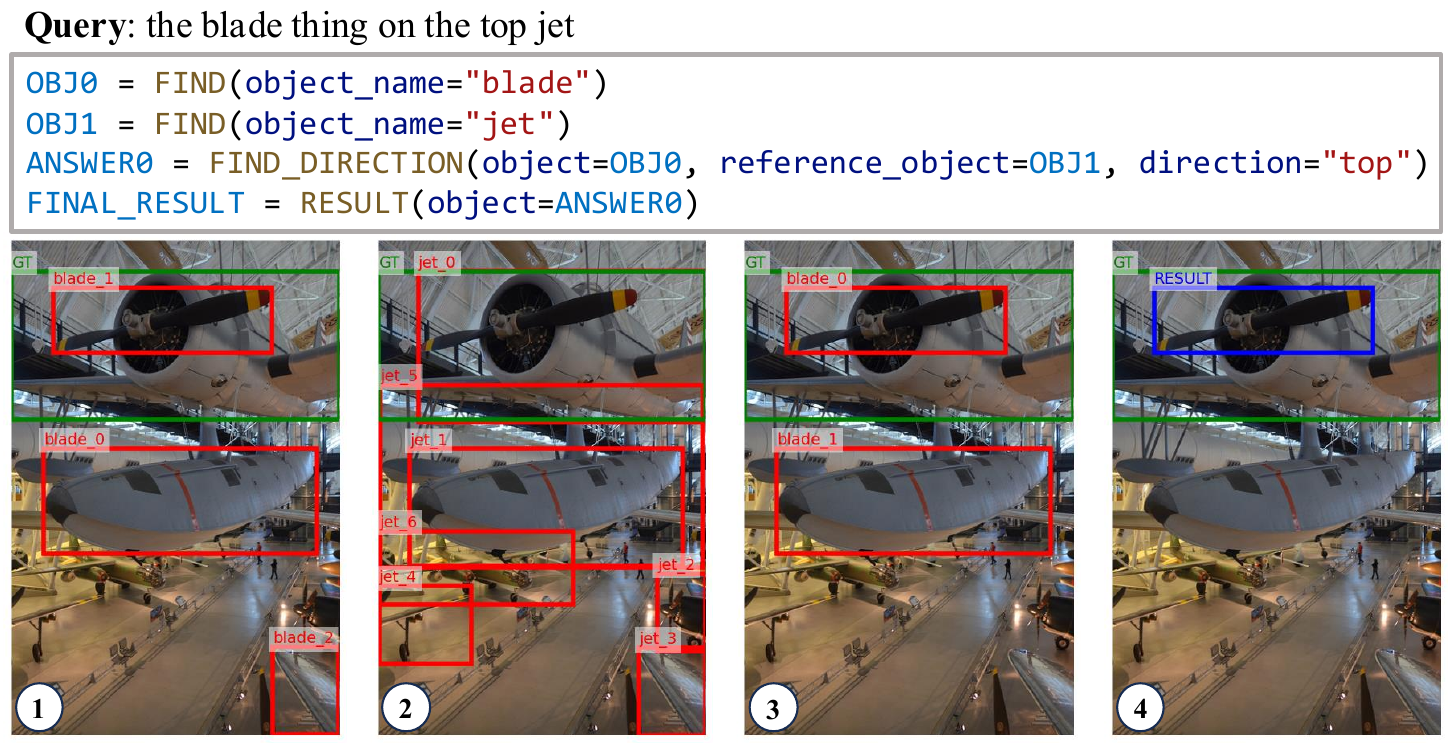}
    \caption{Program generated for the query “the blade thing on the top jet” (top), along with its sequential execution from step 1 to step 4 (bottom).}\label{fig:qual:jet}
\end{figure*}

\begin{figure*}[htb!]
    \centering
    \includegraphics[width=0.75\textwidth]{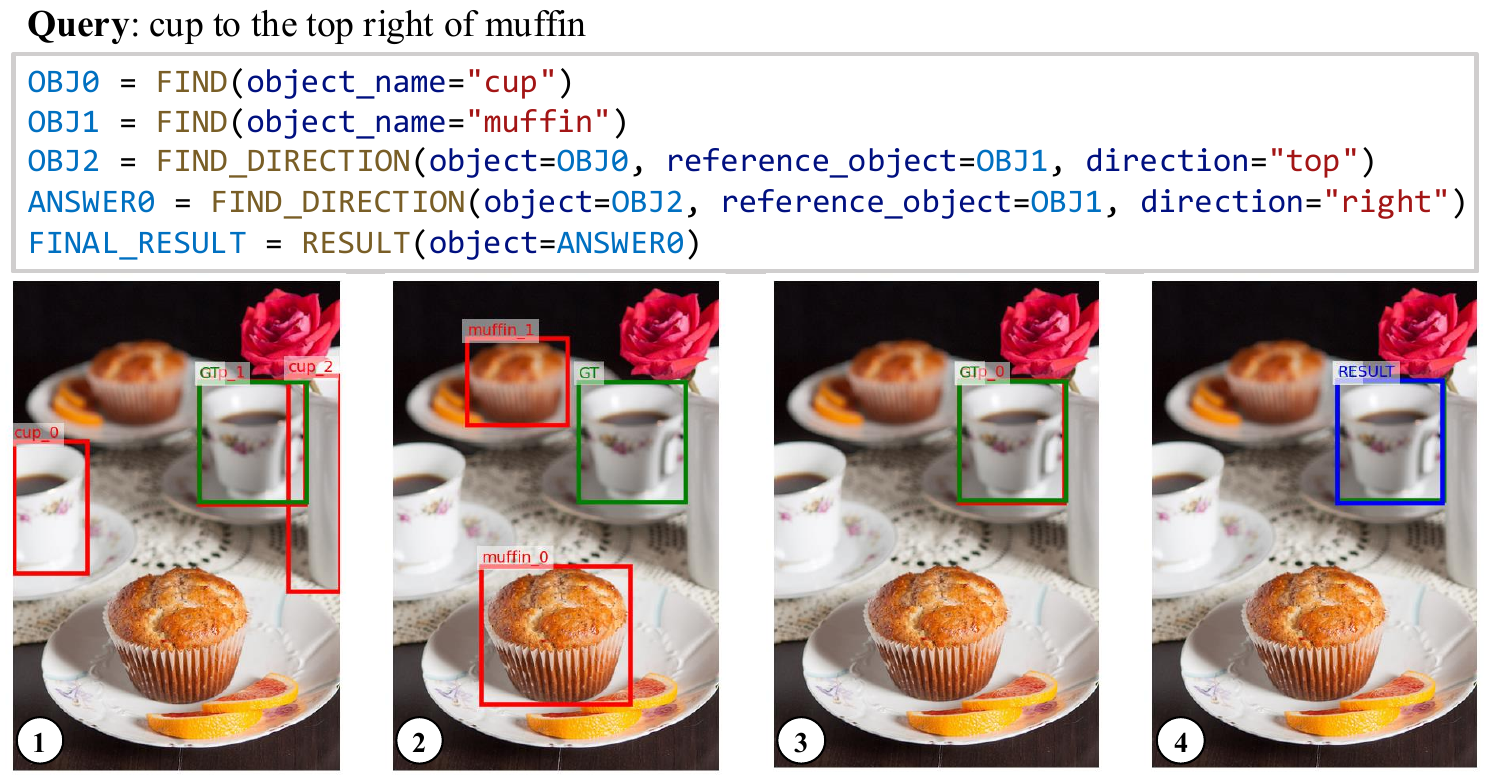}
    \caption{Program generated for the query “cup to the top right of muffin” (top), along with its sequential execution from step 1 to step 4 (bottom).}\label{fig:qual:muffine}
\end{figure*}

\begin{figure*}[htb!]
    \centering
    \includegraphics[width=0.75\textwidth]{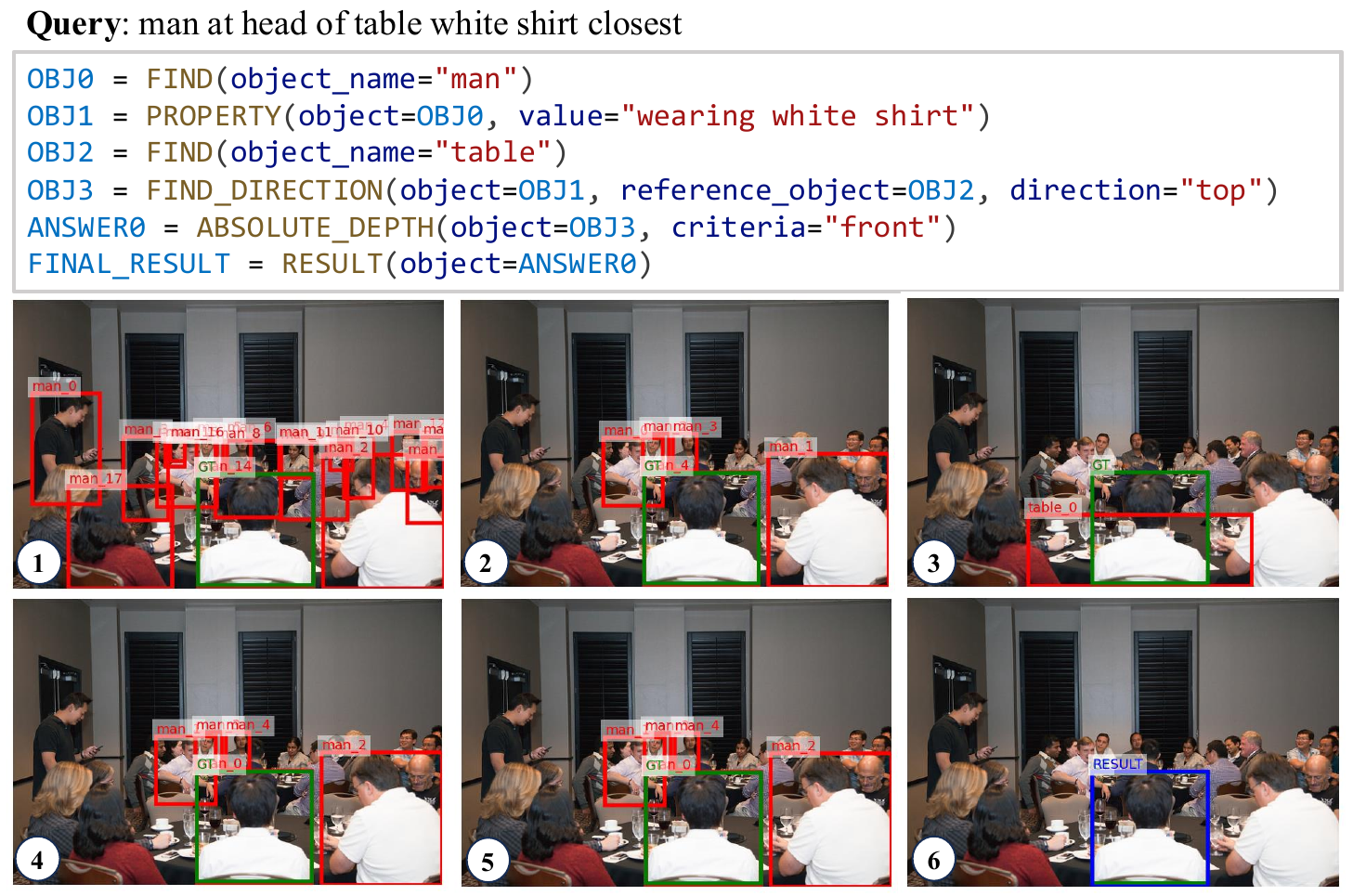}
    \caption{Program generated for the query “man at head of table white shirt closest” (top), along with its sequential execution from step 1 to step 6 (bottom).}\label{fig:qual:person}
\end{figure*}

Figures~\ref{fig:qual:jet},~\ref{fig:qual:muffine},~and~\ref{fig:qual:person} show our program's execution process.

% \subsubsection{Implementation Details}\label{appendix:exp:implementation}

% \begin{table*}[htp!]
%     \centering
%     \begin{tabular}{ccc}
%         \toprule
%         Setting & Parameter & Value \\
%         \midrule
%          \multirow{3}{*}{LLM-related setting} &  temperature & $0.8$ \\
%          & top-$k$  & $50$ \\
%          & top-$p$ & $0.95$ \\
%          \midrule
%          \multirow{2}{*}{General setting } & batch size & $1$ \\
%          & the number of examples (prompts) & $1$ \\
%          \midrule
%          \multirow{6}{*}{MCTS setting} & depth limit & $5$ \\
%          & the number of iterations & $10$ \\
%          & the number of actions & $4$ \\
%          & reward alpha & $0.5$ \\
%          & the number of confidences & $8$ \\
%          & a default value of reward confidence  & $0.8$ \\
%          \midrule
%          \multirow{2}{*}{ToT setting} & beam size & 3 \\
%          & depth limit & $5$ \\
%          \midrule
%          CoT-SC setting & the number of self-consistency & 10 \\ 
%          \bottomrule
%     \end{tabular}
%     \caption{Default hyperparameters for SE, RAP, ToT, and CoT-SC. Parameters are grouped into LLM-related, general, and method-specific settings.}
%     \label{tab:appendix_hyperparam}
% \end{table*}


\end{document}